\documentclass[11pt]{article}
\usepackage{acl07}
\usepackage{times}
\usepackage{latexsym}
\usepackage{color}
\usepackage{stmaryrd}           
\usepackage{amsmath}            
\usepackage[mathscr]{eucal}     
\usepackage{wasysym}
\usepackage{graphicx}
\usepackage{amsthm}             

\makeatletter
\newcommand\figcaption{\def\@captype{figure}\caption}
\newcommand\tabcaption{\def\@captype{table}\caption}
\makeatother

\newcommand\topspace{\rule{0pt}{2.6ex}}
\newcommand\bottomspace{\rule[-1.2ex]{0pt}{0pt}}

\title{Empirical Evaluation of Four Tensor Decomposition Algorithms}

\author{Peter D. Turney \\
Institute for Information Technology \\
National Research Council of Canada \\
M-50 Montreal Road \\
Ottawa, Ontario, Canada \\
K1A 0R6 \\
peter.turney@nrc-cnrc.gc.ca \\ \\
Technical Report ERB-1152, NRC-49877 \\
November 12, 2007}

\begin{document}
\maketitle

\begin{abstract}
Higher-order tensor decompositions are analogous to
the familiar Singular Value Decomposition (SVD), but
they transcend the limitations of matrices (second-order
tensors). SVD is a powerful tool that has achieved impressive
results in information retrieval, collaborative filtering,
computational linguistics, computational vision, and other fields.
However, SVD is limited to two-dimensional arrays of data
(two modes), and many potential applications have three or more
modes, which require higher-order tensor decompositions. This
paper evaluates four algorithms for higher-order tensor
decomposition: Higher-Order Singular Value Decomposition
(HO\nobreakdash-SVD), Higher-Order Orthogonal Iteration (HOOI),
Slice Projection (SP), and Multislice Projection (MP). We measure
the time (elapsed run time), space (RAM and disk space requirements),
and fit (tensor reconstruction accuracy) of the four
algorithms, under a variety of conditions. We find that
standard implementations of HO\nobreakdash-SVD and HOOI do not
scale up to larger tensors, due to increasing RAM requirements.
We recommend HOOI for tensors that are small enough for the
available RAM and MP for larger tensors.
\end{abstract}

\section{Introduction}

Singular Value Decomposition (SVD) is growing increasingly popular as
a tool for the analysis of two-dimensional arrays of data, due to its
success in a wide variety of applications, such as information retrieval
\cite{deerwester1990ils}, collaborative filtering \cite{billsus1998lci},
computational linguistics \cite{schutze1998aws}, computational vision
\cite{brand2002isv}, and genomics \cite{alter2000svd}. SVD is
limited to two-dimensional arrays (matrices or second-order
tensors), but many applications require higher-dimensional arrays, known
as higher-order tensors.

There are several higher-order tensor decompositions, analogous
to SVD, that are able to capture higher-order structure that cannot
be modeled with two dimensions (two modes).
Higher-order generalizations of SVD include Higher-Order Singular Value
Decomposition (HO\nobreakdash-SVD) \cite{delathauwer2000msv}, Tucker
decomposition \cite{tucker1966mnt}, and PARAFAC (parallel factor analysis)
\cite{harshman1970fpp}, which is also known as CANDECOMP (canonical
decomposition) \cite{carroll1970aid}.

Higher-order tensors quickly become unwieldy. The number of elements
in a matrix increases quadratically, as the product of the number of
rows and columns, but the number of elements in a third-order tensor
increases cubically, as a product of the number of rows, columns,
and tubes. Thus there is a need for tensor decomposition algorithms that
can handle large tensors.

In this paper, we evaluate four algorithms for higher-order tensor
decomposition: Higher-Order Singular Value Decomposition
(HO\nobreakdash-SVD) \cite{delathauwer2000msv}, Higher-Order
Orthogonal Iteration (HOOI) \cite{delathauwer2000bra}, Slice
Projection (SP) \cite{wang2005rra}, and Multislice Projection (MP)
(introduced here). Our main concern is the ability of the
four algorithms to scale up to large tensors.

In Section~\ref{sec:apps}, we motivate this work by listing
some of the applications for higher-order tensors. In any field
where SVD has been useful, there is likely to be a third or
fourth mode that has been ignored, because
SVD only handles two modes.

The tensor notation we use in this paper is presented in
Section~\ref{sec:notation}. We follow the notational
conventions of \newcite{kolda2006moh}.

Section~\ref{sec:algorithms} presents the four algorithms,
HO\nobreakdash-SVD, HOOI, SP, and MP. For HO\nobreakdash-SVD and
HOOI, we used the implementations given in the MATLAB Tensor Toolbox
\cite{bader2007emc,bader2007mtt}. For SP and MP, we created
our own MATLAB implementations. Our implementation of MP for
third-order tensors is given in the Appendix.

Section~\ref{sec:experiments} presents our empirical evaluation
of the four tensor decomposition algorithms. In the experiments,
we measure the time (elapsed run time), space (RAM and disk space
requirements), and fit (tensor reconstruction accuracy) of the
four algorithms, under a variety of conditions.

The first group of experiments looks at how the algorithms scale as
the input tensors grow increasingly larger. We test the algorithms
with random sparse third-order tensors as input. HO\nobreakdash-SVD and
HOOI exceed the available RAM when given larger tensors as
input, but SP and MP are able to process large tensors with low RAM
usage and good speed. HOOI provides the best fit, followed by MP,
then SP, and lastly HO\nobreakdash-SVD.

The second group of experiments examines the sensitivity of the fit to
the balance in the ratios of the core sizes (defined in
Section~\ref{sec:notation}). The algorithms are tested with random
sparse third-order tensors as input. In general, the fit of the four
algorithms follows the same pattern as in the first group of experiments
(HOOI gives the best fit, then MP, SP, and HO\nobreakdash-SVD), but we
observe that SP is particularly sensitive to unbalanced ratios of the core
sizes.

The third group explores the fit with varying ratios between the
size of the input tensor and the size of the core tensor. For this group,
we move from third-order tensors to fourth-order tensors. The algorithms
are tested with random fourth-order tensors, with the input tensor size
fixed while the core sizes vary. The fit of the algorithms follows
the same pattern as in the previous two groups of experiments, in spite
of the move to fourth-order tensors.

The final group measures the performance with a real (nonrandom)
tensor that was generated for a task in computational linguistics.
The fit follows the same pattern as in the previous three groups
of experiments. Furthermore, the differences in fit are reflected in
the performance on the given task. This experiment validates the use
of random tensors in the previous three groups of experiments.

We conclude in Section~\ref{sec:conclusions}. There
are tradeoffs in time, space, and fit for the four
algorithms, such that there is no absolute winner
among the four algorithms. The choice will depend
on the time, space, and fit requirements of the given
application. If good fit is the primary concern,
we recommend HOOI for smaller tensors that can fit
in the available RAM, and MP for larger tensors.

\section{Applications}
\label{sec:apps}

A good survey of applications for tensor decompositions for data
analysis is \newcite{acar:umd}, which lists several applications, including
electroencephalogram (EEG) data analysis, spectral analysis of chemical
mixtures, computer vision, and social network analysis. \newcite{kolda2006moh}
also lists various applications, such as psychometrics, image analysis,
graph analysis, and signal processing.

We believe that a natural place to look for applications for
tensor decompositions is wherever SVD has proven useful.
We have grown accustomed to thinking of data in terms of two-dimensional
tables and matrices; in terms of what we can handle with SVD. However,
real applications often have many more modes, which we have been
ignoring.

In information retrieval \cite{deerwester1990ils}, SVD is typically
applied to a \emph{term} $\times$ \emph{document} matrix, where each row
represents a word and each column represents a document in the collection.
An element in the matrix is a weight that represents the importance
of the given word in the given document. SVD smoothes the weights, so
that a document $d$ will have a nonzero weight for a word $w$ if $d$
is similar to other documents that contain the word $w$, even if $d$
does not contain actually contain $w$. Thus a search for $w$ will
return the document $d$, thanks to the smoothing effect of SVD.

To extend the term-document matrix to a third-order tensor, it would
be natural to add information such as author, date of publication,
citations, and venue (e.g., the name of the conference or journal).
For example, \newcite{dunlavy2006maa} used a tensor to combine
information from abstracts, titles, keywords, authors, and citations.
\newcite{chew2007cli} applied a tensor decomposition
to a \emph{term} $\times$ \emph{document} $\times$ \emph{language}
tensor, for cross-language information retrieval.
\newcite{sun2006bsa} analyzed an \emph{author} $\times$ \emph{keyword}
$\times$ \emph{date} tensor.

In collaborative filtering \cite{billsus1998lci}, SVD is usually
applied to a \emph{user} $\times$ \emph{item} matrix, in which each
row represents a person and each column represent an item, such as
a movie or a book. An element in the matrix is a rating by the given
user for the given item. Most of the elements in the
matrix are missing, because each user has only rated a few items.
When a zero element represents a missing rating, SVD can be used to
guess the missing ratings, based on the nonzero elements.

The user-item matrix could be extended to a third-order tensor by
adding a variety of information, such as the words used to describe an
item, the words used to describe the interests of a user, the price
of an item, the geographical location of a user, and the age of a
user. For example, \newcite{mahoney2006tcd} and \newcite{xu2006cas}
applied tensor decompositions to collaborative filtering.

In computational linguistics, SVD is often applied in semantic
space models of word meaning. For example, \newcite{landauer1997sps}
applied SVD to a \emph{word} $\times$ \emph{document} matrix,
achieving human-level scores on multiple-choice synonym questions
from the TOEFL test. \newcite{turney2006ssr} applied SVD to a
\emph{word-pair} $\times$ \emph{pattern} matrix, reaching human-level
scores on multiple-choice analogy questions from the SAT test.

In our recent work, we have begun exploring tensor decompositions
for semantic space models. We are currently developing a
\emph{word} $\times$ \emph{pattern} $\times$ \emph{word} tensor
that can used for both synonyms and analogies. The experiments
in Section~\ref{subsec:real} evaluate the four tensor decomposition
algorithms using this tensor to answer multiple-choice TOEFL
questions.

In computational vision, SVD is often applied to image analysis
\cite{brand2002isv}. To work with the two-mode constraint of SVD,
an image, which is naturally two-dimensional, is mapped to a vector.
For example, in face recognition, SVD is applied to a
\emph{face} $\times$ \emph{image-vector} matrix, in which each
row is a vector that encodes an image of a person's face.
\newcite{wang2005rra} pointed out that this two-mode approach to
image analysis is ignoring essential higher-mode structure in the
data. The experiments in \newcite{wang2005rra} demonstrate that
higher-order tensor decompositions can be much more effective than
SVD.

In summary, wherever SVD has been useful, we expect there are
higher-order modes that have been ignored. With algorithms that
can decompose large tensors, it is no longer necessary to ignore
these modes.

\section{Notation}
\label{sec:notation}

This paper follows the notational conventions of \newcite{kolda2006moh}.
Tensors of order three or higher are represented by bold script letters,
$\boldsymbol{\mathscr{X}}$. Matrices (second-order tensors) are denoted
by bold capital letters, $\mathbf{A}$. Vectors (first-order tensors) are
denoted by bold lowercase letters, $\mathbf{b}$. Scalars (zero-order
tensors) are represented by lowercase italic letters, $i$.

The $i$\nobreakdash-th element in a vector $\mathbf{b}$ is indicated by $b_i$.
The $i$\nobreakdash-th row in a matrix $\mathbf{A}$ is denoted by
$\mathbf{a}_{i:}$, the $j$\nobreakdash-th column is given by
$\mathbf{a}_{:j}$, and the element in row $i$ and column $j$ is represented
by $a_{ij}$.

A third-order tensor $\boldsymbol{\mathscr{X}}$ has rows, columns, and
\emph{tubes}. The element in row $i$, column $j$, and tube $k$ is represented
by $x_{ijk}$. The row vector that contains $x_{ijk}$ is denoted by
$\mathbf{x}_{i:k}$, the column vector is $\mathbf{x}_{:jk}$, and the tube
vector is $\mathbf{x}_{ij:}$. In general, the vectors in a tensor (e.g., the
rows, columns, and tubes in a third-order tensor) are called \emph{fibers}.
There are no special names (beyond rows, columns, and tubes) for fibers
in tensors of order four and higher.

A third-order tensor $\boldsymbol{\mathscr{X}}$ contains matrices,
called \emph{slices}. The horizontal, lateral, and frontal slices of
$\boldsymbol{\mathscr{X}}$ are represented by $\mathbf{X}_{i::}$,
$\mathbf{X}_{:j:}$, and $\mathbf{X}_{::k}$, respectively. The
concept of slices also applies to tensors of order four and higher.

An index $i$ ranges from $1$ to $I$; that is, the upper bound on the
range of an index is given by the uppercase form of the index letter.
Thus the size of a tensor $\boldsymbol{\mathscr{X}}$ is denoted by
uppercase scalars, $I_1 \times I_2 \times I_3$.

There are several kinds of tensor products, but we only need
the $n$\nobreakdash-mode product in this paper.
The $n$\nobreakdash-mode product of a tensor $\boldsymbol{\mathscr{X}}$
and a matrix $\mathbf{A}$ is written as $\boldsymbol{\mathscr{X}}
\times_n \mathbf{A}$. Let $\boldsymbol{\mathscr{X}}$ be of size
$I_1 \times I_2 \times I_3$ and let $\mathbf{A}$ be of size
$J_1 \times J_2$. The $n$\nobreakdash-mode product $\boldsymbol{\mathscr{X}}
\times_n \mathbf{A}$ multiplies fibers in mode $n$ of $\boldsymbol{\mathscr{X}}$
with row vectors in $\mathbf{A}$. Therefore $n$\nobreakdash-mode
multiplication requires that $I_n = J_2$. The result of $\boldsymbol{\mathscr{X}}
\times_n \mathbf{A}$ is a tensor with the same order (same number
of modes) as $\boldsymbol{\mathscr{X}}$, but with the size $I_n$ replaced
by $J_1$. For example, the result of  $\boldsymbol{\mathscr{X}}
\times_3 \mathbf{A}$ is of size $I_1 \times I_2 \times J_1$,
assuming $I_3 = J_2$.

Let $\boldsymbol{\mathscr{X}}$ be an $N$\nobreakdash-th order
tensor of size $I_1 \times \ldots \times I_N$ and let $\mathbf{A}$
be a matrix of size $J \times I_n$. Suppose that $\boldsymbol{\mathscr{Y}}
= \boldsymbol{\mathscr{X}} \times_n \mathbf{A}$. Thus
$\boldsymbol{\mathscr{Y}}$ is of size $I_1 \times \ldots \times I_{n-1}
\times J \times I_{n+1} \times \ldots \times I_N$. The
elements of $\boldsymbol{\mathscr{Y}}$ are defined as follows:

\begin{equation}
y_{i_1 \ldots i_{n-1} j i_{n+1} \ldots i_N} = \sum_{i_n=1}^{I_n}
x_{i_1 \ldots i_N} \; a_{ji_n}
\end{equation}

The transpose of a matrix $\mathbf{A}$ is written as
$\mathbf{A}^{\mathsf{T}}$. We may think of the classical matrix
product $\mathbf{A}\mathbf{B}$ as a special case of
$n$\nobreakdash-mode product:

\begin{equation}
\mathbf{A}\mathbf{B} = \mathbf{A} \times_2 \mathbf{B}^{\mathsf{T}}
= \mathbf{B} \times_1 \mathbf{A}
\end{equation}

\noindent Fibers in mode two of $\mathbf{A}$ (row vectors) are
multiplied with row vectors in $\mathbf{B}^{\mathsf{T}}$, which
are column vectors (mode one) in $\mathbf{B}$.

A tensor $\boldsymbol{\mathscr{X}}$ can be unfolded into a matrix, 
which is called \emph{matricization}. The $n$-mode matricization of
$\boldsymbol{\mathscr{X}}$ is written $\mathbf{X}_{(n)}$,
and is formed by taking the mode $n$ fibers of $\boldsymbol{\mathscr{X}}$
and making them column vectors in $\mathbf{X}_{(n)}$. Let
$\boldsymbol{\mathscr{X}}$ be a tensor of size $I_1 \times I_2 \times I_3$.
The one-mode matricization $\mathbf{X}_{(1)}$ is of size
$I_1 \times (I_2 \; I_3)$:

\begin{eqnarray}
\mathbf{X}_{(1)} &=& [ \, \mathbf{x}_{:11} \; \mathbf{x}_{:21}
\; \ldots \; \mathbf{x}_{:I_{2}I_{3}} \, ] \\
&=& [ \, \mathbf{X}_{::1} \; \mathbf{X}_{::2} \; \ldots \;
\mathbf{X}_{::I_3} \, ]
\end{eqnarray}

\noindent Similarly, the two-mode matricization $\mathbf{X}_{(2)}$ is of
size $I_2 \times (I_1 \; I_3)$:

\begin{eqnarray}
\mathbf{X}_{(2)} &=& [ \, \mathbf{x}_{1:1} \; \mathbf{x}_{2:1}
\; \ldots \; \mathbf{x}_{I_{1}:I_{3}} \, ] \\
&=& [ \, \mathbf{X}_{::1}^{\mathsf{T}} \; \mathbf{X}_{::2}^{\mathsf{T}}
\; \ldots \; \mathbf{X}_{::I_3}^{\mathsf{T}} \, ]
\end{eqnarray}

\noindent Note that $\boldsymbol{\mathscr{Y}} = \boldsymbol{\mathscr{X}}
\times_n \mathbf{A}$ if and only if $\mathbf{Y}_{(n)} = \mathbf{A}
\mathbf{X}_{(n)}$. Thus $n$\nobreakdash-mode matricization relates the
classical matrix product to the $n$\nobreakdash-mode tensor product. In the
special case of second-order tensors, $\mathbf{C}_{(1)} = \mathbf{C}$ and
$\mathbf{C}_{(2)} = \mathbf{C}^{\mathsf{T}}$, hence
$\mathbf{C} = \mathbf{B} \times_1 \mathbf{A}$ if and only if
$\mathbf{C} = \mathbf{A} \mathbf{B}$. Likewise $\mathbf{C} = \mathbf{B}
\times_2 \mathbf{A}$ if and only if $\mathbf{C}^{\mathsf{T}} =
\mathbf{A} \mathbf{B}^{\mathsf{T}}$.

Let $\boldsymbol{\mathscr{G}}$ be a tensor of size
$J_1 \times \ldots \times J_N$. Let $\mathbf{A}^{(1)}, \ldots,
\mathbf{A}^{(N)}$ be matrices such that $\mathbf{A}^{(n)}$ is
of size $I_n \times J_n$. The \emph{Tucker operator} is defined
as follows \cite{kolda2006moh}:

\begin{multline}
\llbracket \boldsymbol{\mathscr{G}} \: ; \mathbf{A}^{(1)},
\mathbf{A}^{(2)}, \ldots, \mathbf{A}^{(N)} \rrbracket \\
\equiv \boldsymbol{\mathscr{G}} \times_1 \mathbf{A}^{(1)}
\times_2 \mathbf{A}^{(2)}
\ldots \times_N \mathbf{A}^{(N)}
\end{multline}

\noindent The resulting tensor is of size $I_1 \times \ldots \times I_N$.

Let $\boldsymbol{\mathscr{X}}$ be a tensor of size
$I_1 \times \ldots \times I_N$. The \emph{Tucker decomposition} of
$\boldsymbol{\mathscr{X}}$ has the following form
\cite{tucker1966mnt,kolda2006moh}:

\begin{equation}
\boldsymbol{\mathscr{X}} \approx \llbracket \boldsymbol{\mathscr{G}} \: ;
\mathbf{A}^{(1)}, \ldots, \mathbf{A}^{(N)} \rrbracket
\end{equation}

\noindent The tensor $\boldsymbol{\mathscr{G}}$ is called the
\emph{core} of the decomposition. Let $\boldsymbol{\mathscr{G}}$
be of size $J_1 \times \ldots \times J_N$.
Each matrix $\mathbf{A}^{(n)}$ is of size $I_n \times J_n$.

The \emph{$n$\nobreakdash-rank} of a tensor $\boldsymbol{\mathscr{X}}$
is the rank of the matrix $\mathbf{X}_{(n)}$. For a second-order
tensor, the one-rank necessarily equals the two-rank, but this is
not true for higher-order tensors. If $J_n$ is equal
to the $n$\nobreakdash-rank of $\boldsymbol{\mathscr{X}}$ for each
$n$, then it is possible for the Tucker decomposition to exactly
equal $\boldsymbol{\mathscr{X}}$. In general, we want $J_n$ less
than the $n$\nobreakdash-rank of $\boldsymbol{\mathscr{X}}$ for each
$n$, yielding a core $\boldsymbol{\mathscr{G}}$ that has lower
$n$\nobreakdash-ranks than $\boldsymbol{\mathscr{X}}$, analogous to a
truncated (thin) SVD. In the special case of a second-order tensor, the
Tucker decomposition $\mathbf{X} \approx \llbracket \mathbf{S} \: ;
\mathbf{U}, \mathbf{V} \rrbracket$ is equivalent to the thin SVD,
$\mathbf{X} \approx \mathbf{U} \mathbf{S} \mathbf{V}^{\mathsf{T}}$.

Suppose we have a tensor $\boldsymbol{\mathscr{X}}$ and
its Tucker decomposition $\hat{\boldsymbol{\mathscr{X}}} = \llbracket
\boldsymbol{\mathscr{G}} \: ;\mathbf{A}^{(1)}, \ldots, \mathbf{A}^{(N)}
\rrbracket$, such that $\boldsymbol{\mathscr{X}} \approx
\hat{\boldsymbol{\mathscr{X}}}$. In the experiments in
Section~\ref{sec:experiments}, we measure the \emph{fit} of the decomposition
$\hat{\boldsymbol{\mathscr{X}}}$ to the original $\boldsymbol{\mathscr{X}}$
as follows:

\begin{equation}
{\rm fit}(\boldsymbol{\mathscr{X}}, \hat{\boldsymbol{\mathscr{X}}}) =
1 - \frac{\left\| \boldsymbol{\mathscr{X}} - \hat{\boldsymbol{\mathscr{X}}}
\right\|_F}{\left\| \boldsymbol{\mathscr{X}} \right\|_F}
\end{equation}

\noindent The Frobenius norm of a tensor $\boldsymbol{\mathscr{X}}$,
$\left\| \boldsymbol{\mathscr{X}} \right\|_F$, is the square root of
the sum of the absolute squares of its elements. The fit is a normalized
measure of the error in reconstructing $\boldsymbol{\mathscr{X}}$ from its
Tucker decomposition $\hat{\boldsymbol{\mathscr{X}}}$. When
$\boldsymbol{\mathscr{X}} = \hat{\boldsymbol{\mathscr{X}}}$,
the fit is 1; otherwise, it is less than 1, and it may be negative
when the fit is particularly poor.

The equivalence between the $n$\nobreakdash-mode tensor product
and the classical matrix product with $n$\nobreakdash-mode matricization
suggests that tensors might be merely a new notation; that
there may be no advantage to using the Tucker decomposition
with tensors instead of using SVD with unfolded (matricized) tensors.
Perhaps the different layers (slices) of the tensor do not actually
interact with each other in any interesting way. This criticism
would be appropriate if the Tucker decomposition used only one
mode, but the decomposition uses all $N$ modes of
$\boldsymbol{\mathscr{X}}$. Because all modes are used, the
layers of the tensor are thoroughly mixed together.

For example, suppose $\boldsymbol{\mathscr{X}} \approx \llbracket
\boldsymbol{\mathscr{G}} \: ; \mathbf{A}, \mathbf{B}, \mathbf{C} \rrbracket$.
Let $\mathbf{X}_{i::}$ be a slice of $\boldsymbol{\mathscr{X}}$.
There is no slice of $\boldsymbol{\mathscr{G}}$, say $\mathbf{G}_{j::}$,
such that we can reconstruct $\mathbf{X}_{i::}$ from $\mathbf{G}_{j::}$,
using $\mathbf{A}$, $\mathbf{B}$, and $\mathbf{C}$. We need all of
$\boldsymbol{\mathscr{G}}$ in order to reconstruct $\mathbf{X}_{i::}$.

All four of the algorithms that we examine in this
paper perform the Tucker decomposition. One reason for our focus
on the Tucker decomposition is that \newcite{bro1998ism} showed that
the Tucker decomposition can be combined with other tensor
decompositions, such as PARAFAC \cite{harshman1970fpp,carroll1970aid}.
In general, algorithms for the Tucker decomposition scale to large tensors
better than most other tensor decomposition algorithms; therefore
it is possible to improve the speed of other algorithms by
first compressing the tensor with the Tucker decomposition.
The slower algorithm (such as PARAFAC) is then applied to the
(relatively small) Tucker core, instead of the whole (large)
input tensor \cite{bro1998ism}. Thus an algorithm that can
perform the Tucker decomposition with large tensors makes it
possible for other kinds of tensor decompositions to be applied
to large tensors.

\section{Algorithms}
\label{sec:algorithms}

This section introduces the four tensor decomposition algorithms.
All four algorithms take as input an arbitrary tensor $\boldsymbol{\mathscr{X}}$
and a desired core size $J_1 \times \ldots \times J_N$
and generate as output a Tucker decomposition $\hat{\boldsymbol{\mathscr{X}}}
= \llbracket \boldsymbol{\mathscr{G}} \: ;\mathbf{A}^{(1)}, \ldots,
\mathbf{A}^{(N)} \rrbracket$, in which the matrices $\mathbf{A}^{(n)}$
are orthonormal.

For HO\nobreakdash-SVD (Higher-Order
Singular Value Decomposition) and HOOI (Higher-Order Orthogonal
Iteration), we show the algorithms specialized for third-order tensors and
generalized for arbitrary tensors. For SP (Slice Projection) and MP
(Multislice Projection), we present the algorithms for third-order and
fourth-order tensors and leave the generalization for arbitrary tensors
as an excercise for the reader. (There is a need for a better notation, to
write the generalization of SP and MP to arbitrary tensors.)

\subsection{Higher-Order SVD}
\label{subsec:HO-SVD}

Figure~\ref{fig:HO-SVD3} presents the HO\nobreakdash-SVD algorithm for
third-order tensors. Figure~\ref{fig:HO-SVDn} gives the
generalization of HO\nobreakdash-SVD for tensors of arbitrary
order \cite{delathauwer2000msv,kolda2006moh}. In the following
experiments, we used the implementation of HO\nobreakdash-SVD in the
MATLAB Tensor Toolbox \cite{bader2007mtt}. HO\nobreakdash-SVD is not
a distinct function in the Toolbox, but it is easily extracted
from the Tucker Alternating Least Squares function, where
it is a component.

\begin{table*}[htbp]
\begin{tabular*}{\textwidth}{@{\extracolsep{\fill}}ll}
\hline
\topspace \textbf{in:} Tensor $\boldsymbol{\mathscr{X}}$ of 
size $I_1 \times I_2 \times I_3$. \\
\textbf{in:} Desired rank of core: $J_1 \times J_2 \times J_3$. \\
\\
$\mathbf{A} \leftarrow J_1$ leading eigenvectors of
$\mathbf{X}_{(1)}\mathbf{X}_{(1)}^\mathsf{T}$ &
-- $\mathbf{X}_{(1)}$ is the unfolding of $\boldsymbol{\mathscr{X}}$
on mode 1 \\
$\mathbf{B} \leftarrow J_2$ leading eigenvectors of
$\mathbf{X}_{(2)}\mathbf{X}_{(2)}^\mathsf{T}$ \\
$\mathbf{C} \leftarrow J_3$ leading eigenvectors of
$\mathbf{X}_{(3)}\mathbf{X}_{(3)}^\mathsf{T}$ \\
\\
$\boldsymbol{\mathscr{G}} \leftarrow \llbracket
\boldsymbol{\mathscr{X}} \: ; \mathbf{A}^{\mathsf{T}}, \mathbf{B}^{\mathsf{T}},
\mathbf{C}^{\mathsf{T}} \rrbracket$ \\
\\
\textbf{out:} $\boldsymbol{\mathscr{G}}$ of size $J_1 \times J_2
\times J_3$ and orthonormal matrices $\mathbf{A}$ of \\
\hspace{0.2in} size $I_1 \times J_1$, $\mathbf{B}$
of size $I_2 \times J_2$, and $\mathbf{C}$ of size $I_3 \times J_3$, \\
\bottomspace \hspace{0.2in} such that $\boldsymbol{\mathscr{X}} \approx
\llbracket \boldsymbol{\mathscr{G}} \: ; \mathbf{A}, \mathbf{B}, \mathbf{C}
\rrbracket$. \\
\hline
\end{tabular*}
\figcaption {Higher-Order Singular Value Decomposition for third-order tensors
\cite{delathauwer2000msv}.}
\label{fig:HO-SVD3}
\end{table*}

\begin{table*}[htbp]
\begin{tabular*}{\textwidth}{@{\extracolsep{\fill}}ll}
\hline
\topspace \textbf{in:} Tensor $\boldsymbol{\mathscr{X}}$ of size $I_1 \times I_2
\times \cdots \times I_N$. \\
\textbf{in:} Desired rank of core: $J_1 \times J_2 \times \cdots \times J_N$. \\
\\
\textbf{for} $n = 1, \ldots, N$ \textbf{do}  \\
\hspace{0.2in} $\mathbf{A}^{(n)} \leftarrow J_n$ leading eigenvectors of
$\mathbf{X}_{(n)}\mathbf{X}_{(n)}^\mathsf{T}$ &
-- $\mathbf{X}_{(n)}$ is the unfolding of $\boldsymbol{\mathscr{X}}$
on mode $n$ \\
\textbf{end for} \\
\\
$\boldsymbol{\mathscr{G}} \leftarrow \llbracket
\boldsymbol{\mathscr{X}} \: ; \mathbf{A}^{(1)\mathsf{T}}, \ldots,
\mathbf{A}^{(N)\mathsf{T}} \rrbracket$ \\
\\
\textbf{out:} $\boldsymbol{\mathscr{G}}$ of size $J_1 \times J_2
\times \cdots \times J_N$ and orthonormal matrices \\
\bottomspace \hspace{0.2in} $\mathbf{A}^{(n)}$ of size $I_n \times J_n$ such that $\boldsymbol{\mathscr{X}}
\approx \llbracket \boldsymbol{\mathscr{G}} \: ; \mathbf{A}^{(1)},
\ldots, \mathbf{A}^{(N)} \rrbracket$. \\
\hline
\end{tabular*}
\figcaption {Higher-Order Singular Value Decomposition for tensors of
arbitrary order \cite{delathauwer2000msv}.}
\label{fig:HO-SVDn}
\end{table*}

HO\nobreakdash-SVD does not attempt to optimize the fit,
${\rm fit}(\boldsymbol{\mathscr{X}}, \hat{\boldsymbol{\mathscr{X}}})$
\cite{kolda2006moh}. That is, HO\nobreakdash-SVD does not produce an optimal
rank\nobreakdash-$J_1,\ldots,J_N$ approximation to $\boldsymbol{\mathscr{X}}$,
because it optimizes for each mode separately, without considering
interactions among the modes. However, we will see in
Section~\ref{sec:experiments} that HO\nobreakdash-SVD often
produces a reasonable approximation, and it is relatively fast.
For more information about HO\nobreakdash-SVD, see
\newcite{delathauwer2000msv}.

\subsection{Higher-Order Orthogonal Iteration}
\label{subsec:hooi}

Figure~\ref{fig:hooi3} presents the HOOI algorithm for
third-order tensors. Figure~\ref{fig:hooin} gives the
generalization of HOOI for tensors of arbitrary order
\cite{delathauwer2000bra,kolda2006moh}. HOOI is implemented
in the MATLAB Tensor Toolbox \cite{bader2007mtt}, in the
Tucker Alternating Least Squares function.

\begin{table*}[htbp]
\begin{tabular*}{\textwidth}{@{\extracolsep{\fill}}ll}
\hline
\topspace \textbf{in:} Tensor $\boldsymbol{\mathscr{X}}$ of 
size $I_1 \times I_2 \times I_3$. \\
\textbf{in:} Desired rank of core: $J_1 \times J_2 \times J_3$. \\
\\
$\mathbf{B} \leftarrow J_2$ leading eigenvectors of
$\mathbf{X}_{(2)}\mathbf{X}_{(2)}^\mathsf{T}$
& -- initialization via HO\nobreakdash-SVD \\
$\mathbf{C} \leftarrow J_3$ leading eigenvectors of
$\mathbf{X}_{(3)}\mathbf{X}_{(3)}^\mathsf{T}$ \\
\\
\textbf{while} not converged \textbf{do} & -- main loop \\
\hspace{0.2in} $\boldsymbol{\mathscr{U}} \leftarrow \llbracket
\boldsymbol{\mathscr{X}} \: ; \mathbf{I}_1, \mathbf{B}^{\mathsf{T}},
\mathbf{C}^{\mathsf{T}} \rrbracket$ \\
\hspace{0.2in} $\mathbf{A} \leftarrow J_1$ leading
eigenvectors of $\mathbf{U}_{(1)}\mathbf{U}_{(1)}^\mathsf{T}$ \\
\hspace{0.2in} $\boldsymbol{\mathscr{V}} \leftarrow \llbracket
\boldsymbol{\mathscr{X}} \: ; \mathbf{A}^{\mathsf{T}}, \mathbf{I}_2,
\mathbf{C}^{\mathsf{T}} \rrbracket$ \\
\hspace{0.2in} $\mathbf{B} \leftarrow J_2$ leading
eigenvectors of $\mathbf{V}_{(2)}\mathbf{V}_{(2)}^\mathsf{T}$ \\
\hspace{0.2in} $\boldsymbol{\mathscr{W}} \leftarrow \llbracket
\boldsymbol{\mathscr{X}} \: ; \mathbf{A}^{\mathsf{T}}, \mathbf{B}^{\mathsf{T}},
\mathbf{I}_3 \rrbracket$ \\
\hspace{0.2in} $\mathbf{C} \leftarrow J_3$ leading
eigenvectors of $\mathbf{W}_{(3)}\mathbf{W}_{(3)}^\mathsf{T}$ \\
\textbf{end while} \\
\\
$\boldsymbol{\mathscr{G}} \leftarrow \llbracket
\boldsymbol{\mathscr{X}} \: ; \mathbf{A}^{\mathsf{T}}, \mathbf{B}^{\mathsf{T}},
\mathbf{C}^{\mathsf{T}} \rrbracket$ \\
\\
\textbf{out:} $\boldsymbol{\mathscr{G}}$ of size $J_1 \times J_2
\times J_3$ and orthonormal matrices $\mathbf{A}$ of \\
\hspace{0.2in} size $I_1 \times J_1$, $\mathbf{B}$
of size $I_2 \times J_2$, and $\mathbf{C}$ of size $I_3 \times J_3$, \\
\bottomspace \hspace{0.2in} such that $\boldsymbol{\mathscr{X}} \approx
\llbracket \boldsymbol{\mathscr{G}} \: ; \mathbf{A}, \mathbf{B}, \mathbf{C}
\rrbracket$. \\
\hline
\end{tabular*}
\figcaption {Higher-Order Orthogonal Iteration for third-order tensors
\cite{delathauwer2000bra,kolda2006moh}. Note that it is not necessary to
initialize $\mathbf{A}$, since the \textbf{while} loop sets $\mathbf{A}$
using $\mathbf{B}$ and $\mathbf{C}$. $\mathbf{I}_i$ is the identity matrix
of size $I_i \times I_i$.}
\label{fig:hooi3}
\end{table*}

\begin{table*}[htbp]
\begin{tabular*}{\textwidth}{@{\extracolsep{\fill}}ll}
\hline
\topspace \textbf{in:} Tensor $\boldsymbol{\mathscr{X}}$ of size $I_1 \times I_2
\times \cdots \times I_N$. \\
\textbf{in:} Desired rank of core: $J_1 \times J_2 \times \cdots \times J_N$. \\
\\
\textbf{for} $n = 2, \ldots, N$ \textbf{do} & -- initialization via HO\nobreakdash-SVD \\
\hspace{0.2in} $\mathbf{A}^{(n)} \leftarrow J_n$ leading eigenvectors of
$\mathbf{X}_{(n)}\mathbf{X}_{(n)}^\mathsf{T}$ \\
\textbf{end for} \\
\\
\textbf{while} not converged \textbf{do} & -- main loop \\
\hspace{0.2in} \textbf{for} $n = 1, \ldots, N$ \textbf{do} \\
\hspace{0.4in} $\boldsymbol{\mathscr{Z}} \leftarrow \llbracket
\boldsymbol{\mathscr{X}} \: ; \mathbf{A}^{(1)\mathsf{T}}, \ldots,
\mathbf{A}^{(n-1)\mathsf{T}}, \mathbf{I}_n, \mathbf{A}^{(n+1)\mathsf{T}},
\ldots, \mathbf{A}^{(N)\mathsf{T}} \rrbracket$ \\
\hspace{0.4in} $\mathbf{A}^{(n)} \leftarrow J_n$ leading
eigenvectors of $\mathbf{Z}_{(n)}\mathbf{Z}_{(n)}^\mathsf{T}$ \\
\hspace{0.2in} \textbf{end for} \\
\textbf{end while} \\
\\
$\boldsymbol{\mathscr{G}} \leftarrow \llbracket
\boldsymbol{\mathscr{X}} \: ; \mathbf{A}^{(1)\mathsf{T}}, \ldots,
\mathbf{A}^{(N)\mathsf{T}} \rrbracket$ \\
\\
\textbf{out:} $\boldsymbol{\mathscr{G}}$ of size $J_1 \times J_2
\times \cdots \times J_N$ and orthonormal matrices \\
\bottomspace \hspace{0.2in} $\mathbf{A}^{(n)}$ of size $I_n \times J_n$ such that $\boldsymbol{\mathscr{X}}
\approx \llbracket \boldsymbol{\mathscr{G}} \: ; \mathbf{A}^{(1)},
\ldots, \mathbf{A}^{(N)} \rrbracket$. \\
\hline
\end{tabular*}
\figcaption {Higher-Order Orthogonal Iteration for tensors of arbitrary order
\cite{delathauwer2000bra,kolda2006moh}. $\mathbf{I}_n$ is the identity matrix
of size $I_n \times I_n$.}
\label{fig:hooin}
\end{table*}

HOOI uses HO\nobreakdash-SVD to initialize the matrices, before entering
the main loop. The implementation in the MATLAB Tensor Toolbox
gives the option of using a random initialization, but initialization
with HO\nobreakdash-SVD usually results in a better fit.

In the main loop, each matrix is optimized individually, while the
other matrices are held fixed. This general method is called
Alternating Least Squares (ALS). HOOI, SP, and MP all use ALS.

The main loop terminates when the change in fit drops below a threshold
or when the number of iterations reaches a maximum, whichever comes first.
To calculate the fit for each iteration, HOOI first calculates
the core $\boldsymbol{\mathscr{G}}$ using $\llbracket
\boldsymbol{\mathscr{X}} \: ; \mathbf{A}^{(1)\mathsf{T}}, \ldots,
\mathbf{A}^{(N)\mathsf{T}} \rrbracket$, and then calculates
$\hat{\boldsymbol{\mathscr{X}}}$ from
$\llbracket \boldsymbol{\mathscr{G}} \: ; \mathbf{A}^{(1)},
\ldots, \mathbf{A}^{(N)} \rrbracket$.
The change in fit is the fit of the Tucker decomposition after the
$t$\nobreakdash-th iteration of the main loop minus the fit from the
previous iteration:

\begin{equation}
\Delta_{\rm fit}(t) = {\rm fit}(\boldsymbol{\mathscr{X}},
\hat{\boldsymbol{\mathscr{X}}}^{(t)}) - {\rm fit}(\boldsymbol{\mathscr{X}},
\hat{\boldsymbol{\mathscr{X}}}^{(t-1)})
\end{equation}

\noindent In the experiments, we set the threshold for $\Delta_{\rm fit}(t)$
at $10^{-4}$ and we set the maximum number of iterations at $50$. (These
are the default values in the MATLAB Tensor Toolbox.) The main loop
usually terminated after half a dozen iterations or fewer, with
$\Delta_{\rm fit}(t)$ less than $10^{-4}$.

As implemented in the MATLAB Tensor Toolbox, calculating the HO-SVD
initialization, the intermediate tensor $\boldsymbol{\mathscr{Z}}$,
and the change in fit, $\Delta_{\rm fit}(t)$, requires bringing the
entire input tensor $\boldsymbol{\mathscr{X}}$
into RAM. Although sparse representations are used, this requirement
limits the size of the tensors that we can process, as we see in
Section~\ref{subsec:sizes}. For more information about
HOOI, see \newcite{delathauwer2000bra} and
\newcite{kolda2006moh}.

\subsection{Slice Projection}
\label{subsec:sp}

Figure~\ref{fig:sp3} presents the SP algorithm for third-order
tensors \cite{wang2005rra}. Although \newcite{wang2005rra} do not
discuss tensors beyond the third-order, the SP algorithm generalizes
to tensors of arbitrary order. For example, Figure~\ref{fig:sp4} shows
SP for fourth-order tensors.

\begin{table*}[htbp]
\begin{tabular*}{\textwidth}{@{\extracolsep{\fill}}ll}
\hline
\topspace \textbf{in:} Tensor $\boldsymbol{\mathscr{X}}$ of
size $I_1 \times I_2 \times I_3$. \\
\textbf{in:} Desired rank of core: $J_1 \times J_2 \times J_3$. \\
\\
$\mathbf{C} \leftarrow$ random matrix of size $I_3 \times J_3$ &
-- normalize columns to unit length \\
\\
\vspace{-0.15in} \textbf{while} not converged \textbf{do} & -- main loop \\
\vspace{0.05in}
\begin{minipage}{2.05in}
\begin{equation*}
\mathbf{M}_{13} \leftarrow \sum^{I_2}_{i=1} \mathbf{X}_{:i:}
\mathbf{C} \mathbf{C}^{\mathsf{T}} \mathbf{X}_{:i:}^{\mathsf{T}}
\end{equation*}
\end{minipage} & -- slices on mode 2 \\
\vspace{-0.15in} \hspace{0.2in} $\mathbf{A} \leftarrow J_1$ leading
eigenvectors of $\mathbf{M}_{13}\mathbf{M}_{13}^\mathsf{T}$ \\
\vspace{0.05in}
\begin{minipage}{2.05in}
\begin{equation*}
\mathbf{M}_{21} \leftarrow \sum^{I_3}_{i=1} \mathbf{X}_{::i}^{\mathsf{T}}
\mathbf{A} \mathbf{A}^{\mathsf{T}} \mathbf{X}_{::i}
\end{equation*}
\end{minipage} & -- slices on mode 3 \\
\vspace{-0.15in} \hspace{0.2in} $\mathbf{B} \leftarrow J_2$ leading
eigenvectors of $\mathbf{M}_{21}\mathbf{M}_{21}^\mathsf{T}$ \\
\vspace{0.05in}
\begin{minipage}{2.05in}
\begin{equation*}
\mathbf{M}_{32} \leftarrow \sum^{I_1}_{i=1} \mathbf{X}_{i::}^{\mathsf{T}}
\mathbf{B} \mathbf{B}^{\mathsf{T}} \mathbf{X}_{i::}
\end{equation*}
\end{minipage} & -- slices on mode 1 \\
\hspace{0.2in} $\mathbf{C} \leftarrow J_3$ leading
eigenvectors of $\mathbf{M}_{32}\mathbf{M}_{32}^\mathsf{T}$ \\
\textbf{end while} \\
\\
$\boldsymbol{\mathscr{G}} \leftarrow \llbracket
\boldsymbol{\mathscr{X}} \: ; \mathbf{A}^{\mathsf{T}}, \mathbf{B}^{\mathsf{T}},
\mathbf{C}^{\mathsf{T}} \rrbracket$ \\
\\
\textbf{out:} $\boldsymbol{\mathscr{G}}$ of size $J_1 \times J_2
\times J_3$ and orthonormal matrices $\mathbf{A}$ of \\
\hspace{0.2in} size $I_1 \times J_1$, $\mathbf{B}$
of size $I_2 \times J_2$, and $\mathbf{C}$ of size $I_3 \times J_3$, \\
\bottomspace \hspace{0.2in} such that $\boldsymbol{\mathscr{X}} \approx
\llbracket \boldsymbol{\mathscr{G}} \: ; \mathbf{A}, \mathbf{B}, \mathbf{C}
\rrbracket$. \\
\hline
\end{tabular*}
\figcaption {Slice Projection for third-order tensors \cite{wang2005rra}.
Note that it is not necessary to initialize $\mathbf{A}$ and $\mathbf{B}$.}
\label{fig:sp3}
\end{table*}

\begin{table*}[htbp]
\begin{tabular*}{\textwidth}{@{\extracolsep{\fill}}ll}
\hline
\topspace \textbf{in:} Tensor $\boldsymbol{\mathscr{X}}$ of
size $I_1 \times I_2 \times I_3 \times I_4$. \\
\textbf{in:} Desired rank of core: $J_1 \times J_2 \times J_3 \times J_4$. \\
\\
$\mathbf{D} \leftarrow$ random matrix of size $I_4 \times J_4$ &
-- normalize columns to unit length \\
\\
\vspace{-0.15in} \textbf{while} not converged \textbf{do} & -- main loop \\
\vspace{0.05in}
\begin{minipage}{2.45in}
\begin{equation*}
\mathbf{M}_{14} \leftarrow \sum^{I_2}_{i=1} \sum^{I_3}_{j=1} \mathbf{X}_{:ij:}
\mathbf{D} \mathbf{D}^{\mathsf{T}} \mathbf{X}_{:ij:}^{\mathsf{T}}
\end{equation*}
\end{minipage} & -- slices on modes 2 and 3 \\
\vspace{-0.15in} \hspace{0.2in} $\mathbf{A} \leftarrow J_1$ leading
eigenvectors of $\mathbf{M}_{14}\mathbf{M}_{14}^\mathsf{T}$ \\
\vspace{0.05in}
\begin{minipage}{2.45in}
\begin{equation*}
\mathbf{M}_{21} \leftarrow \sum^{I_3}_{i=1} \sum^{I_4}_{j=1} \mathbf{X}_{::ij}^{\mathsf{T}}
\mathbf{A} \mathbf{A}^{\mathsf{T}} \mathbf{X}_{::ij}
\end{equation*}
\end{minipage} & -- slices on modes 3 and 4 \\
\vspace{-0.15in} \hspace{0.2in} $\mathbf{B} \leftarrow J_2$ leading
eigenvectors of $\mathbf{M}_{21}\mathbf{M}_{21}^\mathsf{T}$ \\
\vspace{0.05in}
\begin{minipage}{2.45in}
\begin{equation*}
\mathbf{M}_{32} \leftarrow \sum^{I_1}_{i=1} \sum^{I_4}_{j=1} \mathbf{X}_{i::j}^{\mathsf{T}}
\mathbf{B} \mathbf{B}^{\mathsf{T}} \mathbf{X}_{i::j}
\end{equation*}
\end{minipage} & -- slices on modes 1 and 4 \\
\vspace{-0.15in} \hspace{0.2in} $\mathbf{C} \leftarrow J_3$ leading
eigenvectors of $\mathbf{M}_{32}\mathbf{M}_{32}^\mathsf{T}$ \\
\vspace{0.05in}
\begin{minipage}{2.45in}
\begin{equation*}
\mathbf{M}_{43} \leftarrow \sum^{I_1}_{i=1} \sum^{I_2}_{j=1} \mathbf{X}_{ij::}^{\mathsf{T}}
\mathbf{C} \mathbf{C}^{\mathsf{T}} \mathbf{X}_{ij::}
\end{equation*}
\end{minipage} & -- slices on modes 1 and 2 \\
\hspace{0.2in} $\mathbf{D} \leftarrow J_4$ leading
eigenvectors of $\mathbf{M}_{43}\mathbf{M}_{43}^\mathsf{T}$ \\
\textbf{end while} \\
\\
$\boldsymbol{\mathscr{G}} \leftarrow \llbracket
\boldsymbol{\mathscr{X}} \: ; \mathbf{A}^{\mathsf{T}}, \mathbf{B}^{\mathsf{T}},
\mathbf{C}^{\mathsf{T}}, \mathbf{D}^{\mathsf{T}} \rrbracket$ \\
\\
\textbf{out:} $\boldsymbol{\mathscr{G}}$ of size $J_1 \times J_2
\times J_3 \times J_4$ and orthonormal matrices \\
\hspace{0.2in} $\mathbf{A}$ of size $I_1 \times J_1$, $\mathbf{B}$
of size $I_2 \times J_2$, $\mathbf{C}$ of size $I_3 \times J_3$, \\
\bottomspace \hspace{0.2in} and $\mathbf{D}$ of size $I_4 \times J_4$,
such that $\boldsymbol{\mathscr{X}} \approx
\llbracket \boldsymbol{\mathscr{G}} \: ; \mathbf{A}, \mathbf{B}, \mathbf{C},
\mathbf{D} \rrbracket$. \\
\hline
\end{tabular*}
\figcaption {Slice Projection for fourth-order tensors.}
\label{fig:sp4}
\end{table*}

Instead of using HO\nobreakdash-SVD, \newcite{wang2005rra}
initialize SP randomly, to avoid bringing $\boldsymbol{\mathscr{X}}$
into RAM. In Figure~\ref{fig:sp3}, the matrix $\mathbf{C}$
is is filled with random numbers from the uniform distribution over
$[0,1]$ and then the columns are normalized.

Note that HO\nobreakdash-SVD calculates each matrix from
$\boldsymbol{\mathscr{X}}$ alone, whereas HOOI calculates
each matrix from $\boldsymbol{\mathscr{X}}$ and all of the
other matrices. SP lies between HO\nobreakdash-SVD and
HOOI, in that it calculates each matrix from $\boldsymbol{\mathscr{X}}$
and one other matrix.

In the main loop, the input tensor $\boldsymbol{\mathscr{X}}$
is processed one slice at a time, again to avoid bringing the whole
tensor into RAM. Before entering the main loop, the first step
is to calculate the slices and store each slice in a file.
MP requires this same first step. The MATLAB source code for MP,
given in the Appendix, shows how we calculate the slices of
$\boldsymbol{\mathscr{X}}$ without bringing all of
$\boldsymbol{\mathscr{X}}$ into RAM.

Our approach to constructing the slice files assumes that
the input tensor is given in a sparse representation, in which
each nonzero element of the tensor is described by one line
in a file. The description consists of the indices that
specify the location of the nonzero element, followed by
the value of the nonzero element. For example, the
element $x_{ijk}$ of a third-order tensor $\boldsymbol{\mathscr{X}}$
is described as $\left\langle {i,j,k,x_{ijk}} \right\rangle$.
To calculate the $n$\nobreakdash-mode slices, we
first sort the input tensor file by mode $n$. For
example, we generate two-mode slices by sorting on $j$,
the second column of the input file. This puts all of
the elements of an $n$\nobreakdash-mode slice together
consecutively in the file. After sorting on mode $n$, we can
read the sorted file one slice at a time, writing each mode $n$
slice to its own unique file.

To sort the input file, we use the Unix \emph{sort} command.
This command allows the user to specify the amount of RAM
used by the sort buffer. In the following experiments, we
arbitrarily set the buffer to 4 GiB, half the available RAM.
(For Windows, the Unix \emph{sort} command is included in
Cygwin.)

The main loop terminates after a maximum number of iterations
or when the core stops growing, whichever comes first.
The growth of the core is measured as follows:

\begin{equation}
\Delta_{\boldsymbol{\mathscr{G}}}(t) = 1 - \frac{\left\|
\boldsymbol{\mathscr{G}}^{(t-1)} \right\|_F}{\left\|
\boldsymbol{\mathscr{G}}^{(t)} \right\|_F}
\end{equation}

\noindent In this equation, $\boldsymbol{\mathscr{G}}^{(t)}$
is the core after the $t$\nobreakdash-th iteration.
We set the threshold for $\Delta_{\boldsymbol{\mathscr{G}}}(t)$
at $10^{-4}$ and we set the maximum number of iterations at $50$.
The main loop usually terminated after half a dozen iterations
or fewer, with $\Delta_{\boldsymbol{\mathscr{G}}}(t)$ less than
$10^{-4}$.

SP uses $\Delta_{\boldsymbol{\mathscr{G}}}(t)$ as a proxy
for $\Delta_{\rm fit}(t)$, to avoid bringing
$\boldsymbol{\mathscr{X}}$ into RAM. With each iteration,
as the estimates for the matrices improve, the core captures
more of the variation in $\boldsymbol{\mathscr{X}}$, resulting
in growth of the core. It is not necessary to bring
$\boldsymbol{\mathscr{X}}$ into RAM in order to calculate
$\boldsymbol{\mathscr{G}}^{(t)}$; we can calculate
$\boldsymbol{\mathscr{G}}^{(t)}$ one slice at a time, as given
in the Appendix.

For more information about SP, see
\newcite{wang2005rra}. \newcite{wang2005oct} introduced another
low RAM algorithm for higher-order tensors, based on blocks
instead of slices.

\subsection{Multislice Projection}
\label{subsec:sd}

Figure~\ref{fig:mp3} presents the MP algorithm for third-order
tensors. The MP algorithm generalizes to arbitrary order.
Figure~\ref{fig:mp4} shows MP for fourth-order tensors.

\begin{table*}[htbp]
\begin{tabular*}{\textwidth}{@{\extracolsep{\fill}}ll}
\hline
\topspace \textbf{in:} Tensor $\boldsymbol{\mathscr{X}}$ of
size $I_1 \times I_2 \times I_3$. \\
\textbf{in:} Desired rank of core: $J_1 \times J_2 \times J_3$. \\
\vspace{-0.15in} \\
\vspace{0.05in}
\begin{minipage}{2.1in}
\begin{equation*}
\mathbf{M}_{2} \leftarrow \sum^{I_1}_{i=1} \mathbf{X}_{i::}
\mathbf{X}_{i::}^{\mathsf{T}} + \sum^{I_3}_{i=1} \mathbf{X}_{::i}^{\mathsf{T}}
\mathbf{X}_{::i}
\end{equation*}
\end{minipage} & -- pseudo HO\nobreakdash-SVD initialization \\
\vspace{-0.15in} $\mathbf{B} \leftarrow J_2$ leading
eigenvectors of $\mathbf{M}_{2}\mathbf{M}_{2}^\mathsf{T}$ \\
\vspace{0.05in}
\begin{minipage}{2.1in}
\begin{equation*}
\mathbf{M}_{3} \leftarrow \sum^{I_1}_{i=1} \mathbf{X}_{i::}^{\mathsf{T}}
\mathbf{X}_{i::} + \sum^{I_2}_{i=1} \mathbf{X}_{:i:}^{\mathsf{T}}
\mathbf{X}_{:i:}
\end{equation*}
\end{minipage} \\
$\mathbf{C} \leftarrow J_3$ leading
eigenvectors of $\mathbf{M}_{3}\mathbf{M}_{3}^\mathsf{T}$ \\
\\
\vspace{-0.15in} \textbf{while} not converged \textbf{do} & -- main loop \\
\vspace{0.05in}
\begin{minipage}{3.25in}
\begin{equation*}
\mathbf{M}_{1} \leftarrow \sum^{I_3}_{i=1} \mathbf{X}_{::i}
\mathbf{B} \mathbf{B}^{\mathsf{T}} \mathbf{X}_{::i}^{\mathsf{T}}
+ \sum^{I_2}_{i=1} \mathbf{X}_{:i:}
\mathbf{C} \mathbf{C}^{\mathsf{T}} \mathbf{X}_{:i:}^{\mathsf{T}}
\end{equation*}
\end{minipage} & -- slices on modes 2 and 3 \\
\vspace{-0.15in} \hspace{0.2in} $\mathbf{A} \leftarrow J_1$ leading
eigenvectors of $\mathbf{M}_{1}\mathbf{M}_{1}^\mathsf{T}$ \\
\vspace{0.05in}
\begin{minipage}{3.25in}
\begin{equation*}
\mathbf{M}_{2} \leftarrow \sum^{I_3}_{i=1} \mathbf{X}_{::i}^{\mathsf{T}}
\mathbf{A} \mathbf{A}^{\mathsf{T}} \mathbf{X}_{::i}
+ \sum^{I_1}_{i=1} \mathbf{X}_{i::}
\mathbf{C} \mathbf{C}^{\mathsf{T}} \mathbf{X}_{i::}^{\mathsf{T}}
\end{equation*}
\end{minipage} & -- slices on modes 1 and 3 \\
\vspace{-0.15in} \hspace{0.2in} $\mathbf{B} \leftarrow J_2$ leading
eigenvectors of $\mathbf{M}_{2}\mathbf{M}_{2}^\mathsf{T}$ \\
\vspace{0.05in}
\begin{minipage}{3.25in}
\begin{equation*}
\mathbf{M}_{3} \leftarrow \sum^{I_2}_{i=1} \mathbf{X}_{:i:}^{\mathsf{T}}
\mathbf{A} \mathbf{A}^{\mathsf{T}} \mathbf{X}_{:i:}
+ \sum^{I_1}_{i=1} \mathbf{X}_{i::}^{\mathsf{T}}
\mathbf{B} \mathbf{B}^{\mathsf{T}} \mathbf{X}_{i::}
\end{equation*}
\end{minipage} & -- slices on modes 1 and 2 \\
\hspace{0.2in} $\mathbf{C} \leftarrow J_3$ leading
eigenvectors of $\mathbf{M}_{3}\mathbf{M}_{3}^\mathsf{T}$ \\
\textbf{end while} \\
\\
$\boldsymbol{\mathscr{G}} \leftarrow \llbracket
\boldsymbol{\mathscr{X}} \: ; \mathbf{A}^{\mathsf{T}}, \mathbf{B}^{\mathsf{T}},
\mathbf{C}^{\mathsf{T}} \rrbracket$ \\
\\
\textbf{out:} $\boldsymbol{\mathscr{G}}$ of size $J_1 \times J_2
\times J_3$ and orthonormal matrices $\mathbf{A}$ of \\
\hspace{0.2in} size $I_1 \times J_1$, $\mathbf{B}$
of size $I_2 \times J_2$, and $\mathbf{C}$ of size $I_3 \times J_3$, \\
\bottomspace \hspace{0.2in} such that $\boldsymbol{\mathscr{X}} \approx
\llbracket \boldsymbol{\mathscr{G}} \: ; \mathbf{A}, \mathbf{B}, \mathbf{C}
\rrbracket$. \\
\hline
\end{tabular*}
\figcaption {Multislice Projection for third-order tensors. MATLAB source
code for this algorithm is provided in the Appendix.}
\label{fig:mp3}
\end{table*}

\begin{table*}[htbp]
\begin{tabular*}{\textwidth}{@{\extracolsep{\fill}}ll}
\hline
\topspace \textbf{in:} Tensor $\boldsymbol{\mathscr{X}}$ of
size $I_1 \times I_2 \times I_3 \times I_4$. \\
\textbf{in:} Desired rank of core: $J_1 \times J_2 \times J_3 \times J_4$. \\
\vspace{-0.1in} \\
\vspace{0.05in}
\begin{minipage}{4.0in}
\begin{equation*}
\mathbf{M}_{2} \leftarrow \sum^{I_3}_{i=1} \sum^{I_4}_{j=1}
\mathbf{X}_{::ij}^{\mathsf{T}} \mathbf{X}_{::ij}
+ \sum^{I_1}_{i=1} \sum^{I_4}_{j=1} \mathbf{X}_{i::j}
\mathbf{X}_{i::j}^{\mathsf{T}}
+ \sum^{I_1}_{i=1} \sum^{I_3}_{j=1} \mathbf{X}_{i:j:}
\mathbf{X}_{i:j:}^{\mathsf{T}}
\end{equation*}
\end{minipage} \\
\vspace{-0.1in} $\mathbf{B} \leftarrow J_2$ leading
eigenvectors of $\mathbf{M}_{2}\mathbf{M}_{2}^\mathsf{T}$ \\
\vspace{0.05in}
\begin{minipage}{4.0in}
\begin{equation*}
\mathbf{M}_{3} \leftarrow \sum^{I_2}_{i=1} \sum^{I_4}_{j=1}
\mathbf{X}_{:i:j}^{\mathsf{T}} \mathbf{X}_{:i:j}
+ \sum^{I_1}_{i=1} \sum^{I_4}_{j=1} \mathbf{X}_{i::j}^{\mathsf{T}}
\mathbf{X}_{i::j}
+ \sum^{I_1}_{i=1} \sum^{I_2}_{j=1} \mathbf{X}_{ij::}
\mathbf{X}_{ij::}^{\mathsf{T}}
\end{equation*}
\end{minipage} \\
\vspace{-0.1in} $\mathbf{C} \leftarrow J_3$ leading
eigenvectors of $\mathbf{M}_{3}\mathbf{M}_{3}^\mathsf{T}$ \\
\vspace{0.05in}
\begin{minipage}{4.0in}
\begin{equation*}
\mathbf{M}_{4} \leftarrow \sum^{I_2}_{i=1} \sum^{I_3}_{j=1}
\mathbf{X}_{:ij:}^{\mathsf{T}} \mathbf{X}_{:ij:}
+ \sum^{I_1}_{i=1} \sum^{I_3}_{j=1} \mathbf{X}_{i:j:}^{\mathsf{T}}
\mathbf{X}_{i:j:}
+ \sum^{I_1}_{i=1} \sum^{I_2}_{j=1} \mathbf{X}_{ij::}^{\mathsf{T}}
\mathbf{X}_{ij::}
\end{equation*}
\end{minipage} \\
$\mathbf{D} \leftarrow J_4$ leading
eigenvectors of $\mathbf{M}_{4}\mathbf{M}_{4}^\mathsf{T}$ \\
\vspace{0.1in} \\
\vspace{-0.15in} \textbf{while} not converged \textbf{do} \\
\vspace{0.05in}
\begin{minipage}{5.6in}
\begin{equation*}
\mathbf{M}_{1} \leftarrow \sum^{I_3}_{i=1} \sum^{I_4}_{j=1} \mathbf{X}_{::ij}
\mathbf{B} \mathbf{B}^{\mathsf{T}} \mathbf{X}_{::ij}^{\mathsf{T}}
+ \sum^{I_2}_{i=1} \sum^{I_4}_{j=1} \mathbf{X}_{:i:j}
\mathbf{C} \mathbf{C}^{\mathsf{T}} \mathbf{X}_{:i:j}^{\mathsf{T}}
+ \sum^{I_2}_{i=1} \sum^{I_3}_{j=1} \mathbf{X}_{:ij:}
\mathbf{D} \mathbf{D}^{\mathsf{T}} \mathbf{X}_{:ij:}^{\mathsf{T}}
\end{equation*}
\end{minipage} \\
\vspace{-0.1in} \hspace{0.2in} $\mathbf{A} \leftarrow J_1$ leading
eigenvectors of $\mathbf{M}_{1}\mathbf{M}_{1}^\mathsf{T}$ \\
\vspace{0.05in}
\begin{minipage}{5.6in}
\begin{equation*}
\mathbf{M}_{2} \leftarrow \sum^{I_3}_{i=1} \sum^{I_4}_{j=1}
\mathbf{X}_{::ij}^{\mathsf{T}}
\mathbf{A} \mathbf{A}^{\mathsf{T}} \mathbf{X}_{::ij}
+ \sum^{I_1}_{i=1} \sum^{I_4}_{j=1} \mathbf{X}_{i::j}
\mathbf{C} \mathbf{C}^{\mathsf{T}} \mathbf{X}_{i::j}^{\mathsf{T}}
+ \sum^{I_1}_{i=1} \sum^{I_3}_{j=1} \mathbf{X}_{i:j:}
\mathbf{D} \mathbf{D}^{\mathsf{T}} \mathbf{X}_{i:j:}^{\mathsf{T}}
\end{equation*}
\end{minipage} \\
\vspace{-0.1in} \hspace{0.2in} $\mathbf{B} \leftarrow J_2$ leading
eigenvectors of $\mathbf{M}_{2}\mathbf{M}_{2}^\mathsf{T}$ \\
\vspace{0.05in}
\begin{minipage}{5.6in}
\begin{equation*}
\mathbf{M}_{3} \leftarrow \sum^{I_2}_{i=1} \sum^{I_4}_{j=1}
\mathbf{X}_{:i:j}^{\mathsf{T}}
\mathbf{A} \mathbf{A}^{\mathsf{T}} \mathbf{X}_{:i:j}
+ \sum^{I_1}_{i=1} \sum^{I_4}_{j=1} \mathbf{X}_{i::j}^{\mathsf{T}}
\mathbf{B} \mathbf{B}^{\mathsf{T}} \mathbf{X}_{i::j}
+ \sum^{I_1}_{i=1} \sum^{I_2}_{j=1} \mathbf{X}_{ij::}
\mathbf{D} \mathbf{D}^{\mathsf{T}} \mathbf{X}_{ij::}^{\mathsf{T}}
\end{equation*}
\end{minipage} \\
\vspace{-0.1in} \hspace{0.2in} $\mathbf{C} \leftarrow J_3$ leading
eigenvectors of $\mathbf{M}_{3}\mathbf{M}_{3}^\mathsf{T}$ \\
\vspace{0.05in}
\begin{minipage}{5.6in}
\begin{equation*}
\mathbf{M}_{4} \leftarrow \sum^{I_2}_{i=1} \sum^{I_3}_{j=1}
\mathbf{X}_{:ij:}^{\mathsf{T}}
\mathbf{A} \mathbf{A}^{\mathsf{T}} \mathbf{X}_{:ij:}
+ \sum^{I_1}_{i=1} \sum^{I_3}_{j=1} \mathbf{X}_{i:j:}^{\mathsf{T}}
\mathbf{B} \mathbf{B}^{\mathsf{T}} \mathbf{X}_{i:j:}
+ \sum^{I_1}_{i=1} \sum^{I_2}_{j=1} \mathbf{X}_{ij::}^{\mathsf{T}}
\mathbf{C} \mathbf{C}^{\mathsf{T}} \mathbf{X}_{ij::}
\end{equation*}
\end{minipage} \\
\hspace{0.2in} $\mathbf{D} \leftarrow J_4$ leading
eigenvectors of $\mathbf{M}_{4}\mathbf{M}_{4}^\mathsf{T}$ \\
\textbf{end while} \\
\\
$\boldsymbol{\mathscr{G}} \leftarrow \llbracket
\boldsymbol{\mathscr{X}} \: ; \mathbf{A}^{\mathsf{T}}, \mathbf{B}^{\mathsf{T}},
\mathbf{C}^{\mathsf{T}}, \mathbf{D}^{\mathsf{T}} \rrbracket$ \\
\\
\textbf{out:} $\boldsymbol{\mathscr{G}}$ of size $J_1 \times J_2
\times J_3 \times J_4$ and orthonormal matrices \\
\hspace{0.2in} $\mathbf{A}$ of size $I_1 \times J_1$, $\mathbf{B}$
of size $I_2 \times J_2$, $\mathbf{C}$ of size $I_3 \times J_3$, \\
\bottomspace \hspace{0.2in} and $\mathbf{D}$ of size $I_4 \times J_4$,
such that $\boldsymbol{\mathscr{X}} \approx
\llbracket \boldsymbol{\mathscr{G}} \: ; \mathbf{A}, \mathbf{B}, \mathbf{C},
\mathbf{D} \rrbracket$. \\
\hline
\end{tabular*}
\figcaption {Multislice Projection for fourth-order tensors.}
\label{fig:mp4}
\end{table*}

The basic structure of MP is taken from SP, but MP takes three
ideas from HOOI: (1) use HO-SVD to initialize, instead of
random initialization, (2) use fit to determine convergence,
instead of using the growth of the core, (3) use all of the
other matrices to calculate a given matrix, instead of using
only one other matrix. Like SP, MP begins by calculating
all of the slices of the input tensor and storing each slice
in a file. See the Appendix for details.

We call the initialization \emph{pseudo} HO\nobreakdash-SVD
initialization, because it is not exactly HO\nobreakdash-SVD, as
can be seen by comparing the initialization in Figure~\ref{fig:hooi3}
with the initialization in Figure~\ref{fig:mp3}. Note that
$\mathbf{X}_{(2)}$ in Figure~\ref{fig:hooi3} is of size
$I_2 \times (I_1 \; I_3)$, whereas $\mathbf{M}_2$ in
Figure~\ref{fig:mp3} is of size $I_2 \times I_2$, which is
usually much smaller. HO\nobreakdash-SVD brings
the whole tensor into RAM, but pseudo HO\nobreakdash-SVD
processes one slice at a time.

The main loop terminates when the change in fit drops below a threshold
or when the number of iterations reaches a maximum, whichever comes first.
We calculate the fit one slice at a time, as given in the Appendix; it
is not necessary to bring the whole input tensor into RAM in order to
calculate the fit. We set the threshold for $\Delta_{\rm fit}(t)$ at
$10^{-4}$ and we set the maximum number of iterations at $50$. The main
loop usually terminated after half a dozen iterations or fewer, with
$\Delta_{\rm fit}(t)$ less than $10^{-4}$.

The most significant difference between SP and MP is that MP
uses all of the other matrices to calculate a given matrix.
For example, $\mathbf{M}_{13}$ in Figure~\ref{fig:sp3} is based
on $\boldsymbol{\mathscr{X}}$ and $\mathbf{C}$, whereas
the corresponding $\mathbf{M}_1$ in Figure~\ref{fig:mp3} is based
on $\boldsymbol{\mathscr{X}}$, $\mathbf{B}$, and $\mathbf{C}$.
In this respect, MP is like HOOI, as we can see with the
corresponding $\boldsymbol{\mathscr{U}}$ in Figure~\ref{fig:hooi3}.
By slicing on two modes, instead of only one, we improve the
fit of the tensor, as we shall see in the next section.

\section{Experiments}
\label{sec:experiments}

This section presents the four groups of experiments. The hardware
for these experiments was a computer with two dual-core
AMD Opteron 64 processors, 8 GiB of RAM, and a 16 GiB swap file.
The software was 64 bit Suse Linux 10.0, MATLAB R2007a, and MATLAB
Tensor Toolbox Version 2.2 \cite{bader2007mtt}. The algorithms only
used one of the four cores; we did not attempt to perform
parallel processing, although SP and MP could be
parallelized readily.

The input files are plain text files with one line for each
nonzero value in the tensor. Each line consists of integers
that give the location of the nonzero value in the tensor, followed
by a single real number that gives the nonzero value itself.
The input files are in text format, rather than binary format,
in order to facilitate sorting the files.

The output files are binary MATLAB files, containing the
tensor decompositions of the input files. The four algorithms
generate tensor decompositions that are numerically different
but structurally identical. That is, the numerical values are
different, but, for a given input tensor, the four algorithms
generate core tensors and matrices of the same size. Therefore
the output file size does not depend on which algorithm was used.

\subsection{Varying Tensor Sizes}
\label{subsec:sizes}

The goal of this group of experiments was to evaluate the
four algorithms on increasingly larger tensors, to discover
how their performance scales with size. HO\nobreakdash-SVD and HOOI
assume that the input tensor fits in RAM, whereas SP and MP
assume that the input tensor file must be read in blocks.
We expected that HO\nobreakdash-SVD and HOOI would eventually
run out of RAM, but we could not predict precisely how the
four algorithms would scale, in terms of fit, time, and space.

Table~\ref{tab:var-size} summarizes the input test tensors for the
first group of experiments. The test tensors are random sparse
third-order tensors, varying in size from $250^3$ to $2000^3$.
The number of nonzeros in the tensors varies from 1.6 million to
800 million. The nonzero values are random samples from a
uniform distribution between zero and one.

\begin{table*}[htbp]
\begin{tabular*}{\textwidth}{@{\extracolsep{\fill}}cccccc}
\hline
\topspace Input tensor size & Core size & Density & Nonzeros &
Input file & Output file \\
($I_1 \times I_2 \times I_3$) & ($J_1 \times J_2 \times J_3$) &
(\% Nonzero) & (Millions) & (GiB) & (MiB) \\
\hline
\topspace $250 \times 250 \times 250$ & $25 \times 25 \times 25$ &
10 & 1.6 & 0.03 & 0.3 \\
$500 \times 500 \times 500$ & $50 \times 50 \times 50$ &
10 & 12.5 &  0.24 & 1.5 \\
$750 \times 750 \times 750$ & $75 \times 75 \times 75$ &
10 & 42.2 &  0.81 & 4.3 \\
$1000 \times 1000 \times 1000$ & $100 \times 100 \times 100$ &
10 & 100.0 & 1.93 & 9.5 \\
$1250 \times 1250 \times 1250$ & $125 \times 125 \times 125$ &
10 & 195.3 & 3.88 & 17.7 \\
$1500 \times 1500 \times 1500$ & $150 \times 150 \times 150$ &
10 & 337.5 & 6.85 & 29.7 \\
$1750 \times 1750 \times 1750$ & $175 \times 175 \times 175$ &
10 & 535.9 & 11.03 & 46.0 \\
\bottomspace $2000 \times 2000 \times 2000$ & $200 \times 200 \times 200$ &
10 & 800.0 & 16.64 & 67.4 \\
\hline
\end{tabular*}
\caption {Random sparse third-order tensors of varying size.}
\label{tab:var-size}
\end{table*}

Table~\ref{tab:var-size-results} shows the results of the
first group of experiments. HO\nobreakdash-SVD and HOOI
were only able to process the first four tensors, with
sizes from $250^3$ to $1000^3$. The $1000^3$ tensor required
almost 16 GiB of RAM. The next tensor, $1250^3$, required more
RAM than was available (24 GiB; 8 GiB of actual RAM plus a 16
GiB swap file). On the other hand, SP and MP were able to process
all eight tensors, up to $2000^3$. Larger tensors are possible
with SP and MP; the limiting factor becomes run time, rather than
available RAM.

\begin{table*}[htbp]
\begin{tabular*}{\textwidth}{@{\extracolsep{\fill}}cccccccc}
\hline
\topspace Algorithm & Tensor & Nonzeros & Fit &
Run time & Matlab RAM & Sort RAM & Total RAM \\
 & & (Millions) & (\%) & (HH:MM:SS) & (GiB) & (GiB) & (GiB) \\
\hline
\topspace HO\nobreakdash-SVD &  $250^3$ &   1.6 & 3.890 & 00:00:24 &  0.21 & 0.00 &  0.21 \\
HO\nobreakdash-SVD &  $500^3$ &  12.5 & 3.883 & 00:03:44 &  1.96 & 0.00 &  1.96 \\
HO\nobreakdash-SVD &  $750^3$ &  42.2 & 3.880 & 00:14:42 &  6.61 & 0.00 &  6.61 \\
HO\nobreakdash-SVD & $1000^3$ & 100.0 & 3.880 & 01:10:13 & 15.66 & 0.00 & 15.66 \\
\hline
\topspace HOOI &  $250^3$ &   1.6 & 4.053 & 00:01:06 &  0.26 & 0.00 &  0.26 \\
HOOI &  $500^3$ &  12.5 & 3.982 & 00:09:52 &  1.98 & 0.00 &  1.98 \\
HOOI &  $750^3$ &  42.2 & 3.955 & 00:42:45 &  6.65 & 0.00 &  6.65 \\
HOOI & $1000^3$ & 100.0 & 3.942 & 04:01:36 & 15.74 & 0.00 & 15.74 \\
\hline
\topspace SP &  $250^3$ &   1.6 & 3.934 & 00:01:21 & 0.01 & 1.41 & 1.42 \\
SP &  $500^3$ &  12.5 & 3.906 & 00:10:21 & 0.02 & 4.00 & 4.03 \\
SP &  $750^3$ &  42.2 & 3.896 & 00:34:39 & 0.06 & 4.00 & 4.06 \\
SP & $1000^3$ & 100.0 & 3.893 & 01:43:20 & 0.11 & 4.00 & 4.12 \\
SP & $1250^3$ & 195.3 & 3.890 & 03:16:32 & 0.21 & 4.00 & 4.22 \\
SP & $1500^3$ & 337.5 & 3.888 & 06:01:47 & 0.33 & 4.00 & 4.33 \\
SP & $1750^3$ & 535.9 & 3.886 & 09:58:36 & 0.54 & 4.00 & 4.54 \\
SP & $2000^3$ & 800.0 & 3.885 & 15:35:21 & 0.78 & 4.00 & 4.79 \\
\hline
\topspace MP &  $250^3$ &   1.6 & 3.979 & 00:01:45 & 0.01 & 1.41 & 1.42 \\
MP &  $500^3$ &  12.5 & 3.930 & 00:13:55 & 0.03 & 4.00 & 4.03 \\
MP &  $750^3$ &  42.2 & 3.914 & 00:51:33 & 0.06 & 4.00 & 4.07 \\
MP & $1000^3$ & 100.0 & 3.907 & 02:21:30 & 0.12 & 4.00 & 4.12 \\
MP & $1250^3$ & 195.3 & 3.902 & 05:05:11 & 0.22 & 4.00 & 4.23 \\
MP & $1500^3$ & 337.5 & 3.899 & 09:28:49 & 0.37 & 4.00 & 4.37 \\
MP & $1750^3$ & 535.9 & 3.896 & 16:14:01 & 0.56 & 4.00 & 4.56 \\
\bottomspace MP & $2000^3$ & 800.0 & 3.894 & 25:43:17
 & 0.81 & 4.00 & 4.82 \\
\hline
\end{tabular*}
\caption {Performance of the four algorithms with tensors of
varying size.}
\label{tab:var-size-results}
\end{table*}

\begin{figure*}[htbp]
\begin{center}
\includegraphics[scale=0.8]{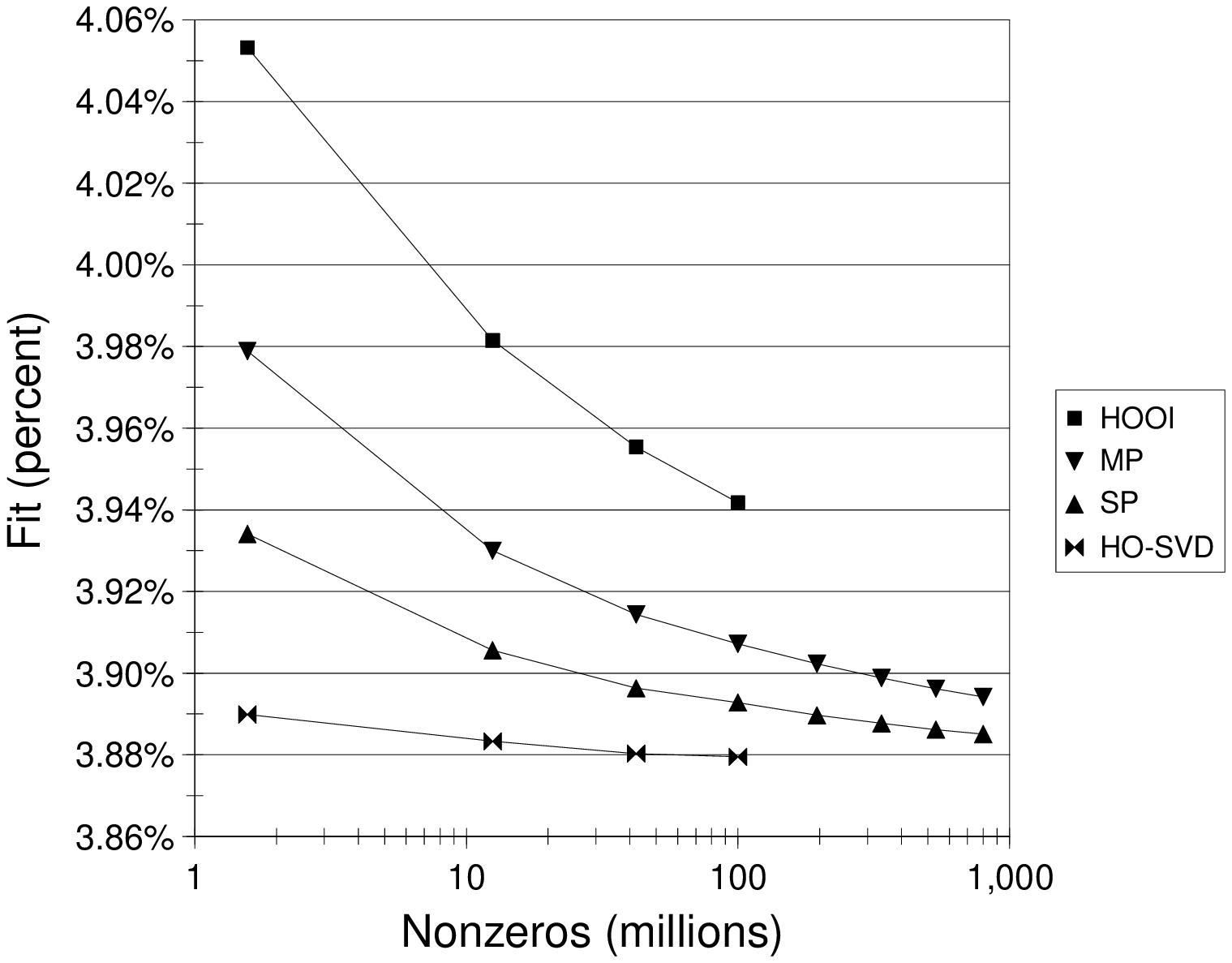}
\caption{The fit of the four algorithms as a function of the
number of nonzeros.}
\label{fig:exper-group1-fit}
\end{center}
\end{figure*}

Figure~\ref{fig:exper-group1-fit} shows the fit of the four algorithms.
HOOI has the best fit, followed by MP, then SP, and finally
HO\nobreakdash-SVD. The curves for HO\nobreakdash-SVD and HOOI
stop at 100 million nonzeros (the $1000^3$ tensor), but it seems
likely that the same trend would continue, if sufficient RAM were
available to apply HO\nobreakdash-SVD and HOOI to the larger tensors.

The fit is somewhat low, at about 4\%, due to the difficulty of
fitting a random tensor with a core size that is 0.1\% of the
size of the input tensor. However, we are interested in the relative
ranking of the four algorithms, rather than the absolute fit.
The results in Section~\ref{subsec:real} show that the ranking we see
here, in Figure~\ref{fig:exper-group1-fit}, is predictive
of the relative performance on a real (nonrandom) task.

\begin{figure*}[htbp]
\begin{center}
\includegraphics[scale=0.8]{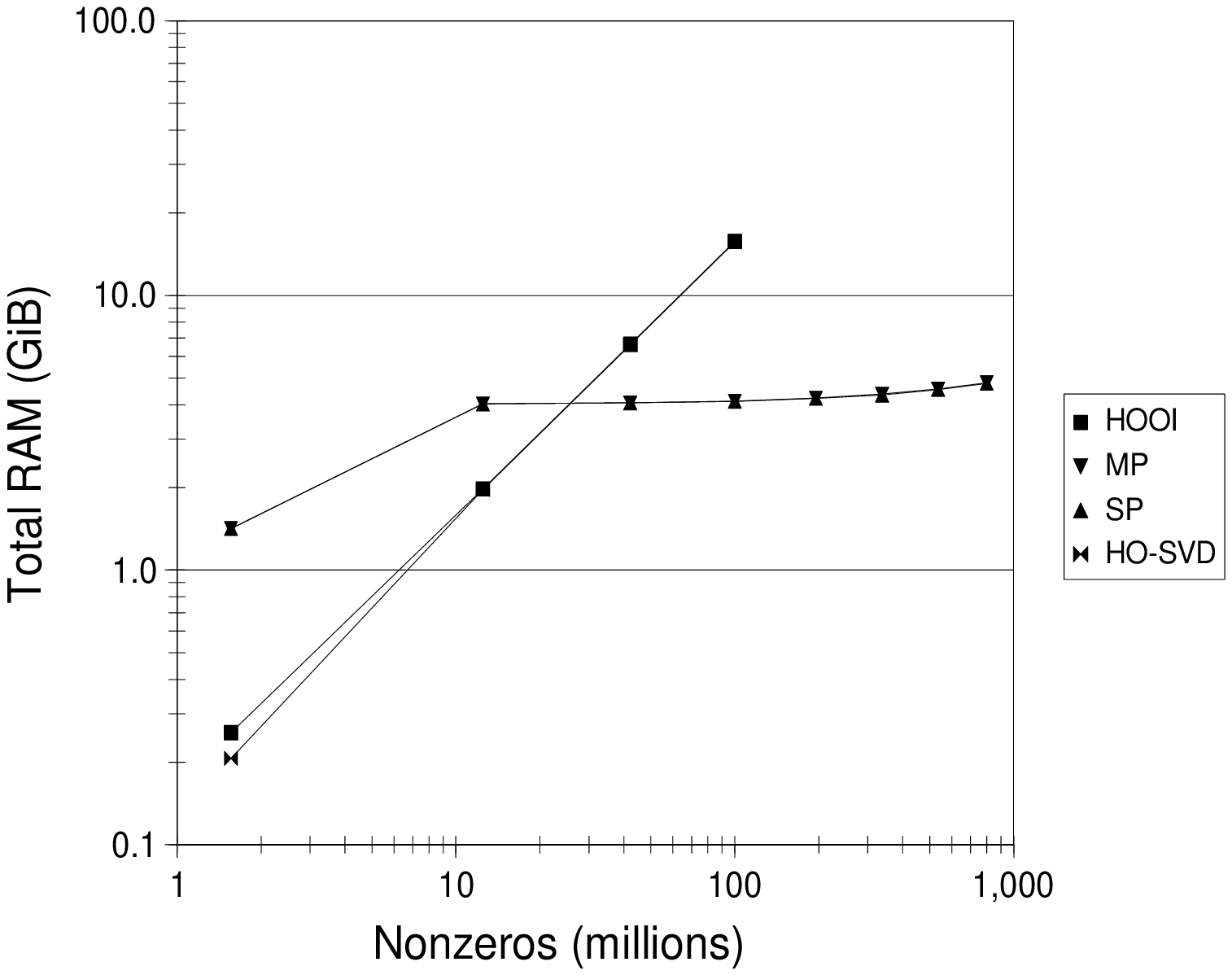}
\caption{The RAM use of the four algorithms as a
function of the number of nonzeros. Note that the
size of the sorting buffer for SP and MP was arbitrarily
set to 4 GiB.}
\label{fig:exper-group1-space}
\end{center}
\end{figure*}

Figure~\ref{fig:exper-group1-space} shows the RAM use of the
algorithms. As we can see in Table~\ref{tab:var-size-results},
there are two components to the RAM use of SP and MP, the
RAM used by \emph{sort} and the RAM used by MATLAB.
We arbitrarily set the sorting buffer to 4 GiB, which
sets an upper bound on the RAM used by \emph{sort}.
A machine with less RAM could use a smaller sorting buffer.
We have not experimented with the buffer size, but we expect
that the buffer could be made much smaller, with only a
slight increase in run time. The growth of the MATLAB
component of RAM use of SP and MP is slow, especially
in comparison to HO\nobreakdash-SVD and HOOI.

\begin{figure*}[htbp]
\begin{center}
\includegraphics[scale=0.8]{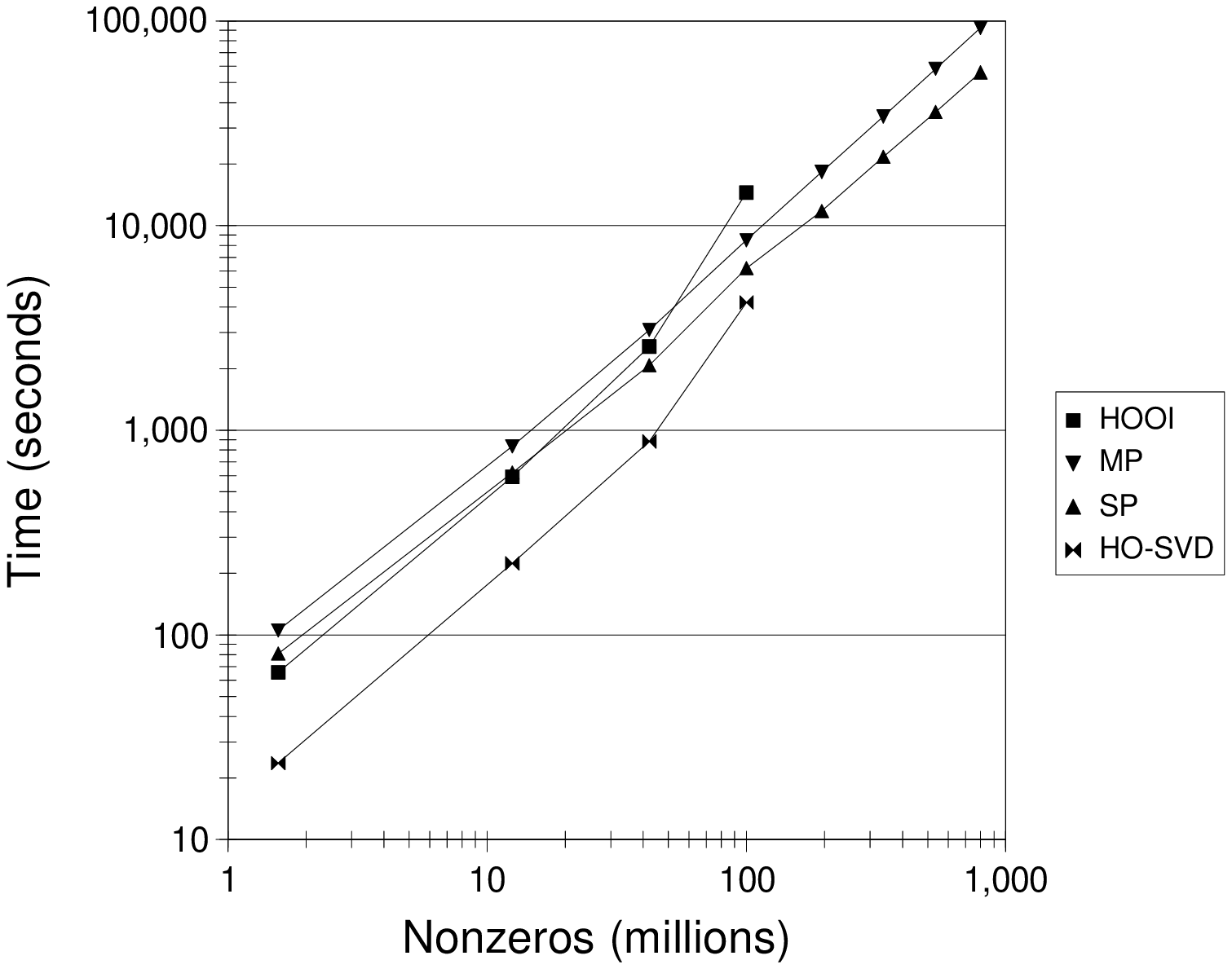}
\caption{The run time of the four algorithms as a function
of the number of nonzeros.}
\label{fig:exper-group1-time}
\end{center}
\end{figure*}

Figure~\ref{fig:exper-group1-time} gives the run time.
For the smallest tensors, SP and MP take longer to run
than HO\nobreakdash-SVD and HOOI, because SP and MP make
more use of files and less use of RAM. With a tensor
size of $1000^3$, both HO\nobreakdash-SVD and HOOI use
up the available hardware RAM (8 GiB) and need to use
the virtual RAM (the 16 GiB swap file), which explains
the sudden upward surge in Figure~\ref{fig:exper-group1-time}
at 100 million nonzeros. In general, the run time
of SP and MP is competitive with HO\nobreakdash-SVD and HOOI.

The results show that SP and MP can handle much larger
tensors than HO\nobreakdash-SVD and HOOI (800 million
nonzeros versus 100 million nonzeros), with only a small
penalty in run time for smaller tensors. However, HOOI
yields a better fit than MP. If fit is important, we
recommend HOOI for smaller tensors and MP for larger
tensors. If speed is more important, we recommend
HO\nobreakdash-SVD for smaller tensors and SP for larger
tensors.

\subsection{Varying Core Size Ratios}
\label{subsec:ratios}

SP is somewhat different from the other three algorithms,
in that it has a kind of asymmetry. Compare $\mathbf{M}_{13}$
in Figure~\ref{fig:sp3} with $\mathbf{M}_1$ in Figure~\ref{fig:mp3}.
We could have used $\mathbf{B}$ instead of $\mathbf{C}$, to
calculate $\mathbf{A}$ in Figure~\ref{fig:sp3}, but we
arbitrarily chose $\mathbf{C}$. We hypothesized that this
asymmetry would make SP sensitive to variation in the
ratios of the core sizes.

\begin{figure*}[htbp]
\begin{center}
\includegraphics[scale=0.8]{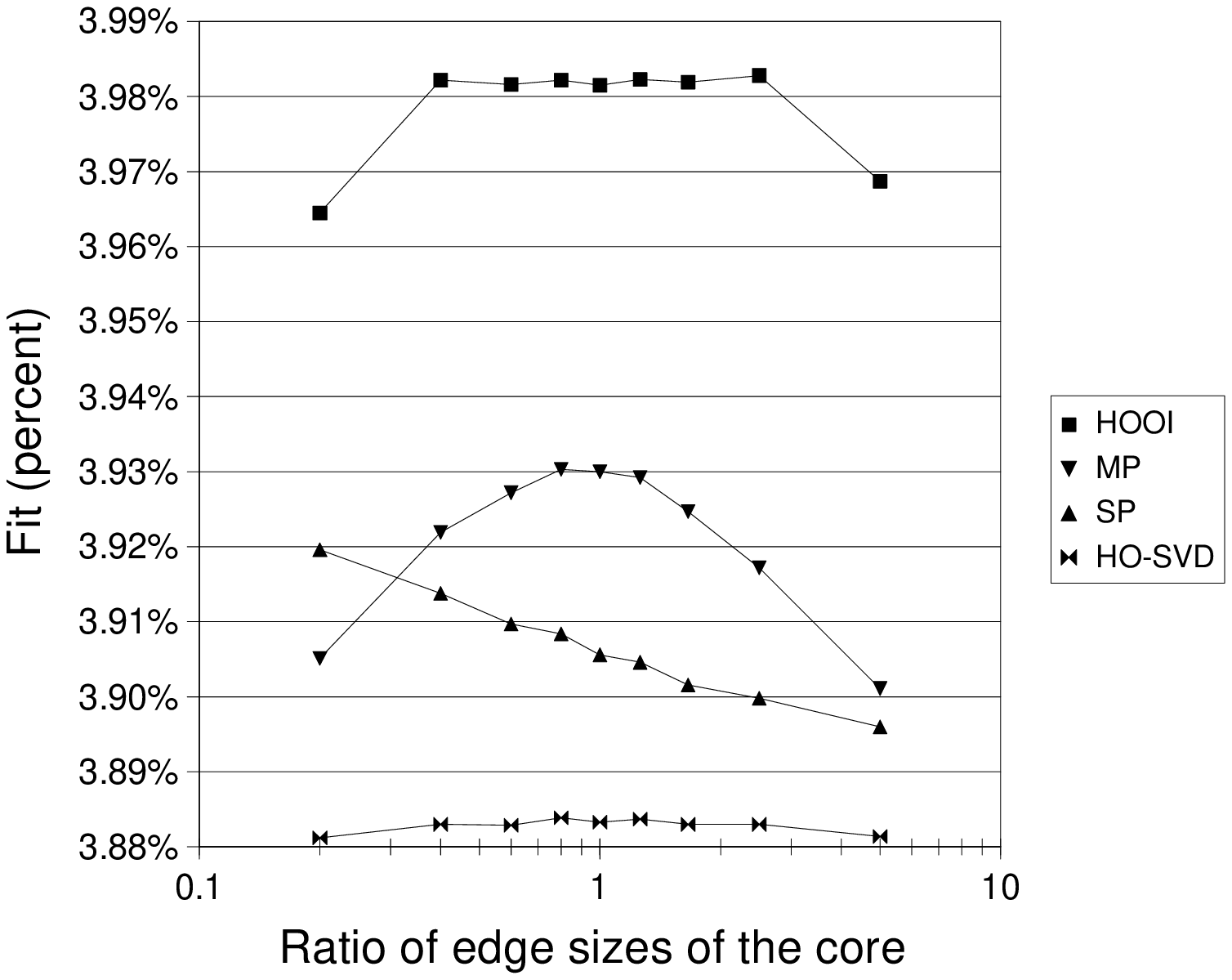}
\caption{The fit of the four algorithms as a function of the
ratios between the sizes of the core in each mode.}
\label{fig:exper-group2-fit}
\end{center}
\end{figure*}

In this group of experiments, we vary the ratios between
the sizes of the core in each mode, as listed in
Table~\ref{tab:var-ratio}. The effect of the ratio on
the performance is shown in Table~\ref{tab:var-ratio-results}.
Figure~\ref{fig:exper-group2-fit} illustrates the effect
of the ratio on the fit. It is clear from the figure that
SP is asymmetrical, whereas HO\nobreakdash-SVD, HOOI, and
MP are symmetrical.

\begin{table*}[htbp]
\begin{tabular*}{\textwidth}{@{\extracolsep{\fill}}ccccccc}
\hline
\topspace Input tensor size & Core size & Ratio & Density & Nonzeros &
Input file & Output file \\
($I_1 \times I_2 \times I_3$) & ($J_1 \times J_2 \times J_3$) &
($J_1 / J_2$ = $J_2 / J_3$) & (\%) & (Millions) & (GiB) & (MiB) \\
\hline
\topspace $500 \times 500 \times 500$ & $250 \times 50 \times 10$
& 5.00 & 10 & 12.5 & 0.24 & 2.05 \\
$500 \times 500 \times 500$ & $125 \times 50 \times 20$
& 2.50 & 10 & 12.5 & 0.24 & 1.63 \\
$500 \times 500 \times 500$ & $83 \times 50 \times 30$
& 1.66 & 10 & 12.5 & 0.24 & 1.51 \\
$500 \times 500 \times 500$ & $63 \times 50 \times 40$
& 1.26 & 10 & 12.5 & 0.24 & 1.48 \\
$500 \times 500 \times 500$ & $50 \times 50 \times 50$
& 1.00 & 10 & 12.5 & 0.24 & 1.46 \\
$500 \times 500 \times 500$ & $40 \times 50 \times 63$
& 0.80 & 10 & 12.5 & 0.24 & 1.48 \\
$500 \times 500 \times 500$ & $30 \times 50 \times 83$
& 0.60 & 10 & 12.5 & 0.24 & 1.51 \\
$500 \times 500 \times 500$ & $20 \times 50 \times 125$
& 0.40 & 10 & 12.5 & 0.24 & 1.63 \\
\bottomspace $500 \times 500 \times 500$ & $10 \times 50 \times 250$
& 0.20 & 10 & 12.5 & 0.24 & 2.05 \\
\hline
\end{tabular*}
\caption {Random sparse third-order tensors with varying ratios
between the sizes of the core in each mode.}
\label{tab:var-ratio}
\end{table*}

\begin{table*}[htbp]
\begin{tabular*}{\textwidth}{@{\extracolsep{\fill}}ccccccc}
\hline
\topspace Algorithm & Ratio & Fit & Run time & Matlab RAM & Sort RAM & Total RAM \\
& ($J_1 / J_2 = J_2 / J_3$) & (\%) & (HH:MM:SS) & (GiB) & (GiB) & (GiB) \\
\hline
\topspace HO\nobreakdash-SVD & 5.00 & 3.881 & 00:06:54 & 2.71 & 0.00 & 2.71 \\
HO\nobreakdash-SVD & 2.50 & 3.883 & 00:04:53 & 1.96 & 0.00 & 1.96 \\
HO\nobreakdash-SVD & 1.66 & 3.883 & 00:04:15 & 1.78 & 0.00 & 1.78 \\
HO\nobreakdash-SVD & 1.26 & 3.884 & 00:03:53 & 1.96 & 0.00 & 1.96 \\
HO\nobreakdash-SVD & 1.00 & 3.883 & 00:03:48 & 1.96 & 0.00 & 1.96 \\
HO\nobreakdash-SVD & 0.80 & 3.884 & 00:03:33 & 1.96 & 0.00 & 1.96 \\
HO\nobreakdash-SVD & 0.60 & 3.883 & 00:03:24 & 1.96 & 0.00 & 1.96 \\
HO\nobreakdash-SVD & 0.40 & 3.883 & 00:03:15 & 1.96 & 0.00 & 1.96 \\
HO\nobreakdash-SVD & 0.20 & 3.881 & 00:03:06 & 1.96 & 0.00 & 1.96 \\
\hline
\topspace HOOI & 5.00 & 3.969 & 00:27:24 & 2.72 & 0.00 & 2.72 \\
HOOI & 2.50 & 3.983 & 00:16:23 & 2.02 & 0.00 & 2.02 \\
HOOI & 1.66 & 3.982 & 00:12:53 & 1.98 & 0.00 & 1.98 \\
HOOI & 1.26 & 3.982 & 00:11:06 & 1.98 & 0.00 & 1.98 \\
HOOI & 1.00 & 3.982 & 00:09:53 & 1.98 & 0.00 & 1.98 \\
HOOI & 0.80 & 3.982 & 00:09:02 & 1.98 & 0.00 & 1.98 \\
HOOI & 0.60 & 3.982 & 00:08:11 & 1.98 & 0.00 & 1.98 \\
HOOI & 0.40 & 3.982 & 00:07:26 & 1.99 & 0.00 & 1.99 \\
HOOI & 0.20 & 3.965 & 00:05:32 & 2.02 & 0.00 & 2.02 \\
\hline
\topspace SP & 5.00 & 3.896 & 00:11:18 & 0.02 & 4.00 & 4.02 \\
SP & 2.50 & 3.900 & 00:09:36 & 0.02 & 4.00 & 4.02 \\
SP & 1.66 & 3.902 & 00:09:30 & 0.02 & 4.00 & 4.02 \\
SP & 1.26 & 3.905 & 00:10:12 & 0.02 & 4.00 & 4.03 \\
SP & 1.00 & 3.906 & 00:10:13 & 0.02 & 4.00 & 4.03 \\
SP & 0.80 & 3.908 & 00:10:12 & 0.03 & 4.00 & 4.03 \\
SP & 0.60 & 3.910 & 00:10:23 & 0.03 & 4.00 & 4.03 \\
SP & 0.40 & 3.914 & 00:10:32 & 0.04 & 4.00 & 4.04 \\
SP & 0.20 & 3.920 & 00:12:43 & 0.06 & 4.00 & 4.07 \\
\hline
\topspace MP & 5.00 & 3.901 & 00:15:01 & 0.02 & 4.00 & 4.03 \\
MP & 2.50 & 3.917 & 00:14:05 & 0.02 & 4.00 & 4.02 \\
MP & 1.66 & 3.925 & 00:13:46 & 0.02 & 4.00 & 4.03 \\
MP & 1.26 & 3.929 & 00:13:47 & 0.02 & 4.00 & 4.03 \\
MP & 1.00 & 3.930 & 00:13:51 & 0.03 & 4.00 & 4.03 \\
MP & 0.80 & 3.930 & 00:13:45 & 0.03 & 4.00 & 4.03 \\
MP & 0.60 & 3.927 & 00:14:17 & 0.03 & 4.00 & 4.04 \\
MP & 0.40 & 3.922 & 00:14:37 & 0.04 & 4.00 & 4.04 \\
\bottomspace MP & 0.20 & 3.905 & 00:16:33 & 0.06 & 4.00 & 4.07 \\
\hline
\end{tabular*}
\caption {Performance of the four algorithms with varying
ratios between the sizes of the core in each mode.}
\label{tab:var-ratio-results}
\end{table*}

This asymmetry of SP might be viewed as a flaw, and thus a
reason for preferring MP over SP, but it could also be seen
as an advantage for SP. In the case where the ratio is 0.2,
SP has a better fit than MP. This suggests that we might
use SP instead of MP when the ratios between the sizes of the
core in each mode are highly skewed; however, we must be
careful to make sure that SP processes the matrices in
the optimal order for the given core sizes.

Note that the relative ranking of the fit of the four
algorithms is the same as in the previous group of experiments
(best fit to worst: HOOI, MP, SP, HO\nobreakdash-SVD), except in
the case of extreme skew. Thus Figure~\ref{fig:exper-group2-fit}
shows the robustness of the relative ranking.

\subsection{Fourth-Order Tensors}
\label{subsec:fourth}

This group of experiments demonstrates that the previous
observations regarding the relative ranking of the fit
also apply to fourth-order tensors. The
experiments also investigate the effect of varying the
size of the core, with a fixed input tensor size.

\begin{figure*}[htbp]
\begin{center}
\includegraphics[scale=0.8]{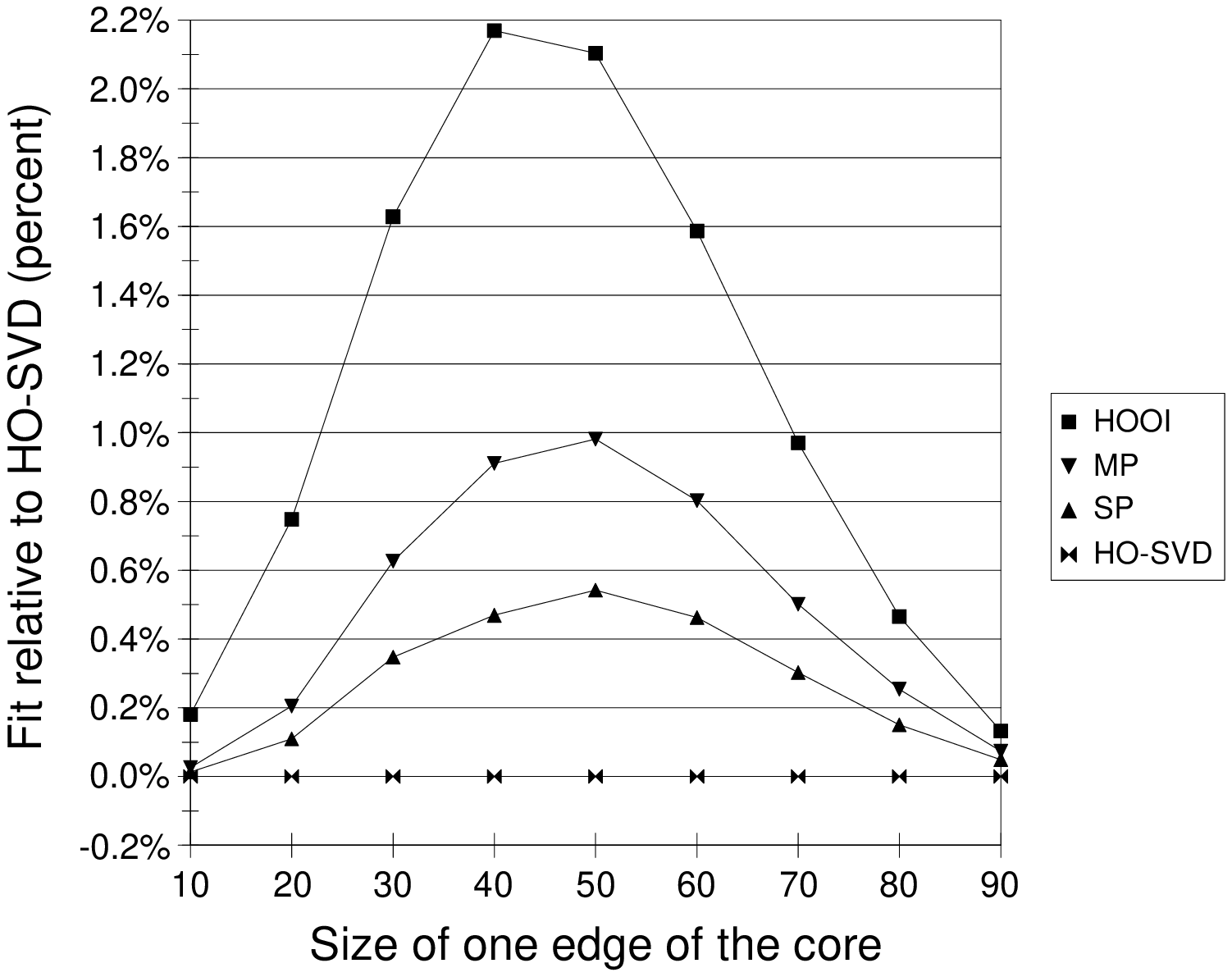}
\caption{The fit of the four algorithms as a function of the
core sizes, given fourth-order tensors.}
\label{fig:exper-group3-fit}
\end{center}
\end{figure*}

Table~\ref{tab:var-core} lists the core sizes that we
investigated. The effect of the core sizes on
the performance is shown in Table~\ref{tab:var-core-results}.
Figure~\ref{fig:exper-group3-fit} shows the impact
of core size on fit.

\begin{table*}[htbp]
\begin{tabular*}{\textwidth}{@{\extracolsep{\fill}}cccccc}
\hline
\topspace Input tensor size & Core size & Density & Nonzeros &
Input file & Output file \\
($I_1 \times I_2 \times I_3 \times I_4$) &
($J_1 \times J_2 \times J_3 \times J_4$) &
(\%) & (Millions) & (MiB) & (MiB) \\
\hline
\topspace $100^4$ & $90^4$ & 10 & 10 & 197.13 & 480.68 \\
$100^4$ & $80^4$ & 10 & 10 & 197.13 & 300.16 \\
$100^4$ & $70^4$ & 10 & 10 & 197.13 & 176.02 \\
$100^4$ & $60^4$ & 10 & 10 & 197.13 & 95.08 \\
$100^4$ & $50^4$ & 10 & 10 & 197.13 & 45.91 \\
$100^4$ & $40^4$ & 10 & 10 & 197.13 & 18.86 \\
$100^4$ & $30^4$ & 10 & 10 & 197.13 & 6.02 \\
$100^4$ & $20^4$ & 10 & 10 & 197.13 & 1.23 \\
\bottomspace $100^4$ & $10^4$ & 10 & 10 & 197.13 & 0.10 \\
\hline
\end{tabular*}
\caption {Random sparse fourth-order tensors with varying core
sizes.}
\label{tab:var-core}
\end{table*}

\begin{table*}[htbp]
\begin{tabular*}{\textwidth}{@{\extracolsep{\fill}}cccccccc}
\hline
\topspace Algorithm & Core Size & Fit & Relative fit
& Run time & Matlab RAM & Sort RAM & Total RAM \\
& & (\%) & (\%) & (HH:MM:SS) & (GiB) & (GiB) & (GiB) \\
\hline
\topspace HO\nobreakdash-SVD & $90^4$ & 44.007 & 0.00 & 00:05:35 & 3.56 & 0.00 & 3.56 \\
HO\nobreakdash-SVD & $80^4$ & 26.477 & 0.00 & 00:05:03 & 3.10 & 0.00 & 3.10 \\
HO\nobreakdash-SVD & $70^4$ & 16.449 & 0.00 & 00:04:20 & 3.02 & 0.00 & 3.02 \\
HO\nobreakdash-SVD & $60^4$ & 10.463 & 0.00 & 00:03:52 & 2.62 & 0.00 & 2.62 \\
HO\nobreakdash-SVD & $50^4$ &  6.988 & 0.00 & 00:03:27 & 2.07 & 0.00 & 2.07 \\
HO\nobreakdash-SVD & $40^4$ &  5.116 & 0.00 & 00:03:13 & 1.89 & 0.00 & 1.89 \\
HO\nobreakdash-SVD & $30^4$ &  4.232 & 0.00 & 00:02:57 & 1.80 & 0.00 & 1.80 \\
HO\nobreakdash-SVD & $20^4$ &  3.903 & 0.00 & 00:02:48 & 1.75 & 0.00 & 1.75 \\
HO\nobreakdash-SVD & $10^4$ &  3.827 & 0.00 & 00:02:34 & 1.80 & 0.00 & 1.80 \\
\hline
\topspace HOOI & $90^4$ & 44.065 & 0.13 & 00:18:35 & 4.32 & 0.00 & 4.32 \\
HOOI & $80^4$ & 26.600 & 0.47 & 00:21:16 & 3.75 & 0.00 & 3.75 \\
HOOI & $70^4$ & 16.609 & 0.97 & 00:17:50 & 3.28 & 0.00 & 3.28 \\
HOOI & $60^4$ & 10.629 & 1.59 & 00:15:00 & 2.88 & 0.00 & 2.88 \\
HOOI & $50^4$ &  7.135 & 2.10 & 00:12:44 & 2.53 & 0.00 & 2.53 \\
HOOI & $40^4$ &  5.227 & 2.17 & 00:10:19 & 2.24 & 0.00 & 2.24 \\
HOOI & $30^4$ &  4.301 & 1.63 & 00:08:42 & 1.97 & 0.00 & 1.97 \\
HOOI & $20^4$ &  3.933 & 0.75 & 00:05:25 & 1.81 & 0.00 & 1.81 \\
HOOI & $10^4$ &  3.834 & 0.18 & 00:04:34 & 1.80 & 0.00 & 1.80 \\
\hline
\topspace SP & $90^4$ & 44.029 & 0.05 & 01:45:07 & 2.19 & 4.00 & 6.19 \\
SP & $80^4$ & 26.517 & 0.15 & 01:31:50 & 1.55 & 4.00 & 5.56 \\
SP & $70^4$ & 16.499 & 0.30 & 01:17:53 & 1.06 & 4.00 & 5.07 \\
SP & $60^4$ & 10.511 & 0.46 & 01:09:49 & 0.44 & 4.00 & 4.44 \\
SP & $50^4$ &  7.026 & 0.54 & 01:04:21 & 0.38 & 4.00 & 4.38 \\
SP & $40^4$ &  5.140 & 0.47 & 01:02:37 & 0.13 & 4.00 & 4.13 \\
SP & $30^4$ &  4.247 & 0.35 & 01:01:09 & 0.07 & 4.00 & 4.08 \\
SP & $20^4$ &  3.908 & 0.11 & 00:59:02 & 0.04 & 4.00 & 4.04 \\
SP & $10^4$ &  3.828 & 0.01 & 00:57:56 & 0.01 & 4.00 & 4.02 \\
\hline
\topspace MP & $90^4$ & 44.039 & 0.07 & 03:16:44 & 2.19 & 4.00 & 6.19 \\
MP & $80^4$ & 26.544 & 0.25 & 02:31:07 & 1.55 & 4.00 & 5.56 \\
MP & $70^4$ & 16.532 & 0.50 & 01:57:17 & 1.06 & 4.00 & 5.07 \\
MP & $60^4$ & 10.547 & 0.80 & 01:36:45 & 0.69 & 4.00 & 4.70 \\
MP & $50^4$ &  7.057 & 0.98 & 01:23:33 & 0.38 & 4.00 & 4.38 \\
MP & $40^4$ &  5.163 & 0.91 & 01:14:23 & 0.17 & 4.00 & 4.18 \\
MP & $30^4$ &  4.259 & 0.63 & 01:07:01 & 0.07 & 4.00 & 4.08 \\
MP & $20^4$ &  3.911 & 0.20 & 01:04:29 & 0.04 & 4.00 & 4.04 \\
\bottomspace MP & $10^4$ &  3.828 & 0.03 & 01:05:26 & 0.01 & 4.00 & 4.02 \\
\hline
\end{tabular*}
\caption {Performance of the four algorithms with fourth-order tensors
and varying core sizes. Relative fit is the percentage increase in fit
relative to HO\nobreakdash-SVD.}
\label{tab:var-core-results}
\end{table*}

The fit varies from about 4\% with a core of $10^4$ to about
44\% with a core of $90^4$. To make the differences among the
algorithms clearer, we normalized the fit by using HO\nobreakdash-SVD
as a baseline. The fit relative to HO\nobreakdash-SVD is defined
as the percentage improvement in the fit of the given algorithm,
compared to the fit of HO\nobreakdash-SVD.

Figure~\ref{fig:exper-group3-fit} shows that the differences among
the four algorithms are largest when the core is about $50^4$;
that is, the size of one mode of the core ($50$) is about half of
the size of one mode of the input tensor ($100$). When the core
is very small or very large, compared to the input tensor,
there is little difference in fit among the algorithms.

The fit follows the same trend here as in the previous two groups
of experiments (best to worst: HOOI, MP, SP, HO\nobreakdash-SVD),
in spite of the switch from third-order tensors to fourth-order
tensors. This further confirms the robustness of the results.

Table~\ref{tab:var-core-results} shows that SP and MP are slow
with fourth-order tensors, compared to HO\nobreakdash-SVD and HOOI.
This is a change from what we observered with third-order tensors,
which did not yield such large differences in run time.
This is because a fourth-order tensor has many more slices than
a third-order tensor with the same number of elements, and each
slice is smaller. There is a much larger overhead associated
with opening and closing many small files, compared to a few
large files. This could be ameliorated by storing several adjacent
slices together in one file, instead of using a separate file
for each slice.

Even with third-order tensors, grouping slices together in one
file would improve the speed of SP and MP. Ideally, the user
would specify the maximum RAM available and SP and MP would
group as many slices together as would fit in the available RAM.

\subsection{Performance with Real Data}
\label{subsec:real}

So far, all our experiments have used random tensors. Our purpose
with this last group of experiments is to show that the previous
observations apply to nonrandom tensors. In particular, the
differences in fit that we have seen so far are somewhat small.
It seems possible that the differences might not matter in a
real application of tensors. This group of experiments shows
that the differences in fit result in differences in performance
on a real task.

The task we examine here is answering multiple-choice synonym
questions from the TOEFL test. This task was first investigated
in \newcite{landauer1997sps}. In ongoing work, we are
exploring the application of third-order tensors to this task,
combining ideas from \newcite{landauer1997sps} and \newcite{turney2006ssr}.

Table~\ref{tab:real-data} describes the input data and the output
tensor decomposition. The first mode of the tensor consists of all of the
391 unique words that occur in the TOEFL questions. The second
mode is a set of 849 words from Basic English, which is an
artificial language that reduces English to a small, easily
learned core vocabulary \cite{ogden1930beg}.
The third mode consists of 1020 patterns that join the words
in the first two modes. These patterns were generated using the
approach of \newcite{turney2006ssr}. The value of an element
in the tensor is derived from the frequency of the corresponding
word pair and pattern in a large corpus.

\begin{table*}[htbp]
\begin{center}
\begin{tabular*}{0.55\textwidth}{llr}
\hline
\topspace Input tensor size & ($I_1 \times I_2 \times I_3$) &
$391 \times 849 \times 1020$ \\
Core size & ($J_1 \times J_2 \times J_3$) & $250 \times 250 \times 250$ \\
Input file & (MiB) &345 \\
Output file & (MiB) & 119 \\
Density & (\% Nonzero) & 5.27 \\
\bottomspace Nonzeros & (Millions) & 18 \\
\hline
\end{tabular*}
\caption {Description of the input data and the output decomposition.}
\label{tab:real-data}
\end{center}
\end{table*}

A TOEFL question consists of a stem word (the target word)
and four choice words. The task is to select the choice word
that is most similar in meaning to the stem word. Our approach
is to measure the similarity of two TOEFL words by the
average similarity of their relations to the Basic English
words.

Let $\boldsymbol{\mathscr{X}}$ be our input tensor.
Suppose we wish to measure the similarity of two TOEFL
words. Let $\mathbf{X}_{i::}$ and $\mathbf{X}_{j::}$
be the slices of $\boldsymbol{\mathscr{X}}$ that correspond
to the two TOEFL words. Each slice gives the weights
for all of the patterns that join the given TOEFL word
to all of the Basic English words. Our measure of
similarity between the TOEFL words is calculated
by comparing the two slices.

Table~\ref{tab:real-results} presents the performance
of the four algorithms. We see that the fit follows
the familiar pattern: HOOI has the best fit, then MP, next
SP, and lastly HO\nobreakdash-SVD. Note that MP and SP
have similar fits. The final column of the table gives
the TOEFL scores for the four algorithms. HOOI has the
best TOEFL score, MP and SP have the same score, and
HO\nobreakdash-SVD has the lowest score. The bottom
row of the table gives the TOEFL score for the raw
input tensor, without the benefit of any smoothing
from the Tucker decomposition. The results validate
the previous experiments with random tensors and illustrate
the value of the Tucker decomposition on a real task.

\begin{table*}[htbp]
\begin{tabular*}{\textwidth}{@{\extracolsep{\fill}}cccccccc}
\hline
\topspace Algorithm & Fit & Relative fit & Run time & Matlab RAM & Sort RAM & Total RAM & TOEFL \\
& (\%) & (\%) & (HH:MM:SS) & (GiB) & (GiB) & (GiB) & (\%) \\
\hline
\topspace HO\nobreakdash-SVD & 21.716 & 0.00 & 00:10:28 & 5.29 & 0.00 & 5.29 & 80.00 \\
HOOI & 22.597 & 4.05 & 00:56:08 & 5.77 & 0.00 & 5.77 & 83.75 \\
SP & 22.321 & 2.78 & 00:30:02 & 0.33 & 4.00 & 4.33 & 81.25 \\
MP & 22.371 & 3.01 & 00:43:52 & 0.33 & 4.00 & 4.34 & 81.25 \\
\bottomspace Raw tensor & - & - & - & - & - & - & 67.50 \\
\hline
\end{tabular*}
\caption {Performance of the four algorithms with actual data.
Relative fit is the percentage increase in fit relative to HO\nobreakdash-SVD.}
\label{tab:real-results}
\end{table*}

\section{Conclusions}
\label{sec:conclusions}

The Tucker decomposition has been with us since 1966, but it
seems that it has only recently started to become popular.
We believe that this is because only recently has computer
hardware reached the point where large tensor decompositions
are becoming feasible.

SVD started to attract interest in the field of information
retrieval when it was applied to ``problems of reasonable size
(1000-2000 document abstracts; and 5000-7000 index terms)''
\cite{deerwester1990ils}. In collaborative filtering, SVD
attracted interest when it achieved good results on the
Netflix Prize, a dataset with a sparse matrix of 17,000
movies rated by 500,000 users. In realistic applications,
size matters. The MATLAB Tensor Toolbox
\cite{bader2007emc,bader2007mtt} has done much to make
tensor decompositions more accessible and easier to
experiment with, but, as we have seen here, RAM requirements
become problematic with tensors larger than $1000^3$.

The aim of this paper has been to empirically evaluate
four tensor decompositions, to study their fit and their time and
space requirements. Our primary concern was the ability of
the algorithms to scale up to large tensors. The implementations
of HO\nobreakdash-SVD and HOOI, taken from the MATLAB Tensor Toolbox,
assumed that the input tensor could fit in RAM, which limited
them to tensors of size $1000^3$. On the other hand, SP and MP
were able to process tensors of size $2000^3$, with eight times
more elements.

The experiments in Section~\ref{subsec:real} suggest that
the differences in fit among the four algorithms correspond
to differences in performance on real tasks. It seems likely
that good fit will be important for many applications; therefore,
we recommend HOOI for those tensors that can fit in the available
RAM, and MP for larger tensors.

\section*{Acknowledgements}

Thanks to Brandyn Webb, Tamara Kolda, and Hongcheng Wang for helpful comments.
Thanks to Tamara Kolda and Brett Bader for the MATLAB Tensor Toolbox.

\bibliographystyle{acl}
\bibliography{NRC-49877}

\onecolumn

\section*{Appendix: MATLAB Source for Multislice Projection}

{\small
\begin{verbatim}
-------------------------------------------------------------------------------

function fit = multislice(data_dir,sparse_file,tucker_file,I,J)
%MULTISLICE is a low RAM Tucker decomposition
%
%   Peter Turney
%   October 26, 2007
%
%   Copyright 2007, National Research Council of Canada
%
%   This program is free software: you can redistribute it and/or modify
%   it under the terms of the GNU General Public License as published by
%   the Free Software Foundation, either version 3 of the License, or
%   (at your option) any later version.
%
%   This program is distributed in the hope that it will be useful,
%   but WITHOUT ANY WARRANTY; without even the implied warranty of
%   MERCHANTABILITY or FITNESS FOR A PARTICULAR PURPOSE.  See the
%   GNU General Public License for more details.
%
%   You should have received a copy of the GNU General Public License
%   along with this program.  If not, see <http://www.gnu.org/licenses/>.
%
%%  set parameters
%
fprintf('MULTISLICE is running ...\n');
%
maxloops = 50;           % maximum number of iterations
eigopts.disp = 0;        % suppress messages from eigs()
minfitchange = 1e-4;     % minimum change in fit of tensor
%
%%  make slices of input data file
%
fprintf('   preparing slices\n');
%
mode1_dir = 'slice1';
mode2_dir = 'slice2';
mode3_dir = 'slice3';
%
slice(data_dir,sparse_file,mode1_dir,1,I);
slice(data_dir,sparse_file,mode2_dir,2,I);
slice(data_dir,sparse_file,mode3_dir,3,I);
%
%%  pseudo HO-SVD initialization
%
%   initialize B
%
M2 = zeros(I(2),I(2));
for i = 1:I(3)
    X3_slice = load_slice(data_dir,mode3_dir,i);
    M2 = M2 + (X3_slice' * X3_slice);
end
for i = 1:I(1)
    X1_slice = load_slice(data_dir,mode1_dir,i);
    M2 = M2 + (X1_slice * X1_slice');
end
[B,D] = eigs(M2*M2',J(2),'lm',eigopts);
%
%   initialize C
%
M3 = zeros(I(3),I(3));
for i = 1:I(1)
    X1_slice = load_slice(data_dir,mode1_dir,i);
    M3 = M3 + (X1_slice' * X1_slice);
end
for i = 1:I(2)
    X2_slice = load_slice(data_dir,mode2_dir,i);
    M3 = M3 + (X2_slice' * X2_slice);
end
[C,D] = eigs(M3*M3',J(3),'lm',eigopts);
%
%%  main loop
%
old_fit = 0;
%
fprintf('   entering main loop of MULTISLICE\n');
%
for loop_num = 1:maxloops
    %
    %   update A
    %
    M1 = zeros(I(1),I(1));
    for i = 1:I(2)
        X2_slice = load_slice(data_dir,mode2_dir,i);
        M1 = M1 + ((X2_slice * C) * (C' * X2_slice'));
    end
    for i = 1:I(3)
        X3_slice = load_slice(data_dir,mode3_dir,i);
        M1 = M1 + ((X3_slice * B) * (B' * X3_slice'));
    end
    [A,D] = eigs(M1*M1',J(1),'lm',eigopts);
    %
    %   update B
    %
    M2 = zeros(I(2),I(2));
    for i = 1:I(3)
        X3_slice = load_slice(data_dir,mode3_dir,i);
        M2 = M2 + ((X3_slice' * A) * (A' * X3_slice));
    end
    for i = 1:I(1)
        X1_slice = load_slice(data_dir,mode1_dir,i);
        M2 = M2 + ((X1_slice * C) * (C' * X1_slice'));
    end
    [B,D] = eigs(M2*M2',J(2),'lm',eigopts);
    %
    %   update C
    %
    M3 = zeros(I(3),I(3));
    for i = 1:I(1)
        X1_slice = load_slice(data_dir,mode1_dir,i);
        M3 = M3 + ((X1_slice' * B) * (B' * X1_slice));
    end
    for i = 1:I(2)
        X2_slice = load_slice(data_dir,mode2_dir,i);
        M3 = M3 + ((X2_slice' * A) * (A' * X2_slice));
    end
    [C,D] = eigs(M3*M3',J(3),'lm',eigopts);
    %
    %   build the core
    %
    G = zeros(I(1)*J(2)*J(3),1);
    G = reshape(G,[I(1) J(2) J(3)]);
    for i = 1:I(1)
        X1_slice = load_slice(data_dir,mode1_dir,i);
        G(i,:,:) = B' * X1_slice * C;
    end
    G = reshape(G,[I(1) (J(2)*J(3))]);
    G = A' * G;
    G = reshape(G,[J(1) J(2) J(3)]);
    %
    %   measure fit
    %
    normX = 0;
    sqerr = 0;
    for i = 1:I(1)
        X1_slice = load_slice(data_dir,mode1_dir,i);
        X1_approx = reshape(G,[J(1) (J(2)*J(3))]);
        X1_approx = A(i,:) * X1_approx;
        X1_approx = reshape(X1_approx,[J(2) J(3)]);
        X1_approx = B * X1_approx * C';
        sqerr = sqerr + norm(X1_slice-X1_approx,'fro')^2;
        normX = normX + norm(X1_slice,'fro')^2;
    end
    fit = 1 - sqrt(sqerr) / sqrt(normX);
    %
    fprintf('   loop %d: fit = %f\n', loop_num, fit);
    %
    %   stop if fit is not increasing fast enough
    %
    if ((fit - old_fit) < minfitchange)
        break;
    end
    %
    old_fit = fit;
    %
end
%
fprintf('   total loops = %d\n', loop_num);
%
%%  save tensor
%
output_file = [data_dir, '/', tucker_file];
save(output_file,'G','A','B','C');
%
fprintf('   tucker tensor is in %s\n',tucker_file);
%
fprintf('MULTISLICE is done\n');
%

-------------------------------------------------------------------------------

function slice(data_dir,sparse_file,mode_slice_dir,mode,I)
%SLICE chops a tensor into slices along the given mode
%
%   Peter Turney
%   October 20, 2007
%
%   Copyright 2007, National Research Council of Canada
%
%%  initialize
%
%   set the secondary modes
%
if (mode == 1)
    r_mode = 2;
    c_mode = 3;
elseif (mode == 2)
    r_mode = 1;
    c_mode = 3;
else
    r_mode = 1;
    c_mode = 2;
end
%
%   get sizes
%
Ns = I(mode);        % number of slices
Nr = I(r_mode);      % number of rows in each slice
Nc = I(c_mode);      % number of columns in each slice
%
%%  sort the index
%
fprintf('SLICE is running ...\n');
%
%   file names
%
sub_dir     = [data_dir, '/', mode_slice_dir];
sorted_file = [sub_dir, '/', 'sorted.txt'];
%
%   make sure the directories exist
%
if (isdir(data_dir) == 0)
    mkdir(data_dir);
end
if (isdir(sub_dir) == 0)
    mkdir(sub_dir);
end
%
%   sort
%
sort_index(data_dir,sparse_file,mode_slice_dir,mode);
%
%%  count nonzeros in each slice
%
fprintf('   counting nonzeros in each slice for mode %d\n',mode);
%
%   vector for storing nonzero count
%
nonzeros = zeros(Ns,1);
%
%   read sorted file in blocks
%
%   - read in blocks because file may be too big to fit in RAM
%   - textscan will create one cell for each field
%   - each cell will contain a column vector of the values in
%     the given field
%   - the number of elements in each column vector is the number
%     of lines that were read
%
desired_lines = 100000;
actual_lines  = desired_lines;
%
sorted_file_id = fopen(sorted_file, 'r');
while (actual_lines > 0)
    block = textscan(sorted_file_id,'%d %d %d %*f',desired_lines);
    mode_subs = block{mode};
    actual_lines = size(mode_subs,1);
    for i = 1:actual_lines
        nonzeros(mode_subs(i)) = nonzeros(mode_subs(i)) + 1;
    end
end
fclose(sorted_file_id);
%
%%  make slices
%
fprintf('   saving slices for mode %d\n',mode);
%
sorted_file_id = fopen(sorted_file, 'r');
for i = 1:Ns
    slice_file = sprintf('%s/slice%d.mat', sub_dir, i);
    nonz = nonzeros(i);
    block = textscan(sorted_file_id,'%d %d %d %f',nonz);
    slice_rows = double(block{r_mode});
    slice_cols = double(block{c_mode});
    slice_vals = block{4};
    slice = sparse(slice_rows,slice_cols,slice_vals,Nr,Nc,nonz);
    save(slice_file,'slice');
end
fclose(sorted_file_id);
%
fprintf('SLICE is done\n');
%

-------------------------------------------------------------------------------

function sort_index(data_dir,sparse_file,mode_slice_dir,mode)
%SORT_INDEX sorts a sparse tensor index file along the given mode
%
%   Peter Turney
%   October 20, 2007
%
%   Copyright 2007, National Research Council of Canada
%
%%  sort the index
%
fprintf('SORT_INDEX is running ...\n');
%
%   file names
%
input_file  = [data_dir, '/', sparse_file];
sub_dir     = [data_dir, '/', mode_slice_dir];
sorted_file = [sub_dir, '/', 'sorted.txt'];
%
%   call Unix 'sort' command
%
%   -n = numerical sorting
%   -k = key to sort on
%   -s = stable sorting
%   -S = memory for sorting buffer
%   -o = output file
%
%   - the 'sort' command is a standard part of Unix and Linux
%   - if you are running Windows, you can get 'sort' by
%     installing Cygwin
%   - the sort buffer is set here to 1 GiB; you can set it
%     to some other value, based on how much RAM you have
%
command = sprintf('sort -n -s -S 1G -k %d,%d -o %s %s', ...
    mode, mode, sorted_file, input_file);
%
fprintf('   calling Unix sort for mode %d\n', mode);
unix(command);
%
fprintf('SORT_INDEX is done\n');
%

-------------------------------------------------------------------------------

function slice = load_slice(data_dir,mode_dir,i)
%LOAD_SLICE loads a sparse slice file
%
%   Peter Turney
%   October 20, 2007
%
%   Copyright 2007, National Research Council of Canada
%
%   file name
%
slice_file = sprintf('%s/%s/slice%d.mat', data_dir, mode_dir, i);
%
%   load the file
%
data = load(slice_file);
%
%   return the slice
%
slice = data.slice;
%

-------------------------------------------------------------------------------

function test
%TEST illustrates how to use multislice.m
%
%   Peter Turney
%   October 26, 2007
%
%   Copyright 2007, National Research Council of Canada
%
%   test multislice.m
%
%   set random seed for repeatable experiments
%
rand('seed',5678);
%
%   set parameters
%
I = [100 110 120];      % input sparse tensor size
J = [10 11 12];         % desired core tensor size
density = 0.1;          % percent nonzero
%
data_dir    = 'test';         % directory for storing tensor
sparse_file = 'spten.txt';    % file for storing raw data tensor
tucker_file = 'tucker.mat';   % file for storing Tucker tensor
%
%   make a sparse random tensor and store it in a file
%
sparse_random_tensor(data_dir,sparse_file,I,density);
%
%   call multislice
%
tic;
fit = multislice(data_dir,sparse_file,tucker_file,I,J);
time = toc;
%
%   show results
%
fprintf('\n');
fprintf('Multislice:\n');
fprintf('I        = [%d %d %d]\n', I(1), I(2), I(3));
fprintf('J        = [%d %d %d]\n', J(1), J(2), J(3));
fprintf('density  = %f\n', density);
fprintf('fit      = %f\n', fit);
fprintf('time     = %.1f\n', time);
fprintf('\n');
%

-------------------------------------------------------------------------------

function sparse_random_tensor(data_dir,sparse_file,I,density)
%SPARSE_RANDOM_TENSOR makes a sparse uniformly distributed random tensor
%
%   Peter Turney
%   October 20, 2007
%
%   Copyright 2007, National Research Council of Canada
%
%   assume a third-order tensor is desired
%
%%  initialize
%
fprintf('SPARSE_RANDOM_TENSOR is running ...\n');
%
%   make sure the directory exists
%
if (isdir(data_dir) == 0)
    mkdir(data_dir);
end
%
file_name = [data_dir, '/', sparse_file];
%
fprintf('   generating tensor of size %d x %d x %d with density %f\n', ...
    I(1), I(2), I(3), density);
%
%%  main loop
%
file_id = fopen(file_name, 'w');
fprintf('   slice: ');
for i1 = 1:I(1)
    fprintf('%d ',i1);       % show progress
    if ((mod(i1,10) == 0) && (i1 ~= I(1)))
        fprintf('\n   ');    % time for new line
    end
    for i2 = 1:I(2)
        for i3 = 1:I(3)
            if (rand < density)
                fprintf(file_id,'%d %d %d %f\n',i1,i2,i3,rand);
            end
        end
    end
end
fprintf('\n');
fclose(file_id);
%
fprintf('SPARSE_RANDOM_TENSOR is done\n');
%

-------------------------------------------------------------------------------
\end{verbatim}
}

\end{document}